\documentclass[letterpaper]{article} % DO NOT CHANGE THIS
\usepackage{aaai2026}  % DO NOT CHANGE THIS
\usepackage{times}  % DO NOT CHANGE THIS
\usepackage{helvet}  % DO NOT CHANGE THIS
\usepackage{courier}  % DO NOT CHANGE THIS
\usepackage[hyphens]{url}  % DO NOT CHANGE THIS
\usepackage{graphicx} % DO NOT CHANGE THIS
\usepackage{natbib}  % DO NOT CHANGE THIS AND DO NOT ADD ANY OPTIONS TO IT
\usepackage{caption} % DO NOT CHANGE THIS AND DO NOT ADD ANY OPTIONS TO IT
\usepackage{multirow}
\usepackage{aaai2026}
\usepackage{tikz}
\usepackage{enumitem}
\usepackage{pgfplots}
\pgfplotsset{compat=1.18}
\usepackage{amsmath}
\usepackage{cleveref}
\usepackage{adjustbox}
\usetikzlibrary{positioning, shapes.geometric, 3d, arrows.meta, positioning, decorations.pathreplacing, calc}
\usepgfplotslibrary{colormaps,fillbetween}

\usepackage[utf8]{inputenc} % allow utf-8 input
\usepackage[T1]{fontenc}    % use 8-bit T1 fonts
\usepackage{snu-causal-2024}
\usepackage{url}            % simple URL typesetting
\usepackage{booktabs}       % professional-quality tables
\usepackage{amsfonts}       % blackboard math symbols
\usepackage{nicefrac}       % compact symbols for 1/2, etc.
\usepackage{microtype}      % microtypography
\usepackage{xcolor}         % colors
\usepackage{longtable}
\usepackage{array}
\usepackage{ragged2e}

\usepackage{newfloat}
\usepackage{listings}
\DeclareCaptionStyle{ruled}{labelfont=normalfont,labelsep=colon,strut=off} % DO NOT CHANGE THIS
\floatstyle{ruled}
\newfloat{listing}{tb}{lst}{}
\floatname{listing}{Listing}
\title{Mitigating Length Bias in RLHF through a Causal Lens}
\author{
    Hyeonji Kim\textsuperscript{\rm 1},
    Sujeong Oh\textsuperscript{\rm 1},
    Sanghack Lee\textsuperscript{\rm 1}\thanks{Corresponding author.}
}
\affiliations{
    \textsuperscript{\rm 1}Graduate School of Data Science, Seoul National University \\
}

\usepackage{bibentry}
\begin{document}

\maketitle

\begin{abstract}
Reinforcement learning from human feedback (RLHF) is widely used to align large language models (LLMs) with human preferences. However, RLHF-trained reward models often exhibit length bias---a systematic tendency to favor longer responses by conflating verbosity with quality. We propose a causal framework for analyzing and mitigating length bias in RLHF reward modeling. Central to our approach is a counterfactual data augmentation method that generates response pairs designed to isolate content quality from verbosity. These counterfactual examples are then used to train the reward model, enabling it to assess responses based on content quality independently of verbosity. Specifically, we construct (1) length-divergent pairs with similar content and (2) content-divergent pairs of similar length. Empirical evaluations show that our method reduces length bias in reward assignment and leads to more concise, content-focused outputs from the policy model. These findings demonstrate that the proposed approach effectively reduces length bias and improves the robustness and content sensitivity of reward modeling in RLHF pipelines.

% \begingroup
% \renewcommand\thefootnote{}
% \footnotetext{Extended version including full technical appendices is available at: \url{https://arxiv.org/abs/XXXX.XXXXX}.}
% \endgroup

\end{abstract}

\section{Introduction}
\label{sec:introduction}

%% 전체적으로 말 다듬고 수정하기
Large language models (LLMs) have demonstrated remarkable performance across a wide range of natural language tasks~\citep{brown2020language, liang2023holisticevaluationlanguagemodels, chowdhery2023palm}. Reinforcement learning from human feedback (RLHF)~\citep{ziegler2020, stiennon2022} has become the dominant approach for aligning LLM behavior with human preferences~\citep{ouyang2022traininglanguagemodelsfollow}. Despite its success, RLHF often inherits and amplifies systematic biases inherently present in human preference data, with \textit{length bias} being one of the most persistent issues~\citep{ouyang2022traininglanguagemodelsfollow, shen2023loose, saito2023verbositybiaspreferencelabeling}. Length bias refers to the tendency of reward models to assign higher scores to longer responses, even when informativeness and relevance are comparable or worse. This bias can significantly distort model behavior and user experience~\citep{singhal2024a}.

Recent studies have empirically demonstrated that both models and human annotators are susceptible to verbosity bias~\citep{saito2023verbositybiaspreferencelabeling, shen2023loose}, often preferring longer responses even when content is held constant. When such preferences are implicitly encoded into reward models, RLHF-trained LLMs tend to prioritize verbosity over clarity, often resulting in unnecessarily long and less effective outputs that may degrade user experience. This phenomenon may arise because reward models leverage spurious correlations in data that fail to capture the true quality of the output~\citep{stiennon2022, singhal2024a, huang2024post}.

Several methods have been proposed to mitigate length bias. For instance,~\citet{chen2024odin, wang2025rewardhackingcausalrewards} operate on learned representations by regularizing the reward model, and~\citet{liu2025rrmrobustrewardmodel, cai2025} generate randomized or loosely controlled response pairs to reduce sensitivity to length. However, these approaches often lack the ability to explicitly disentangle verbosity from semantic quality, still leaving the reward model vulnerable to spurious correlations between response length and the reward.

To guide the reward model toward learning preferences based on content quality rather than surface length features, we adopt a causal perspective on length bias. Without such a framework, it is difficult to separate genuine effects from misleading patterns. For example, although there is a strong correlation between a country's per capita chocolate consumption and its number of Nobel laureates, both are influenced by a third factor such as national wealth. This classic example highlights the risk of relying solely on observational correlations without causal reasoning.

To address this, we propose a \textit{counterfactual data augmentation} framework that enables reward models to disentangle content quality from response length. While counterfactual data augmentation has been used to mitigate spurious correlations in classification tasks \citep{kaushik2019learning}, our work extends this causal idea to the RLHF reward-modeling by asking: “How would the reward change if the same content were expressed more concisely?” To answer this, we generate two types of counterfactual preference pairs: (1) semantically equivalent responses of different lengths, and (2) semantically different responses of similar lengths. These comparisons isolate the effects of content and verbosity, guiding the reward model to develop preferences based on semantic quality rather than length. Our contributions are:
\begin{itemize}
    \item 
    We identify a key limitation of existing approaches to mitigating length bias: their limited ability to disentangle verbosity from semantic quality due to reliance on spurious correlation. Our analysis shows that, without explicit interventions, current methods often conflate response length with informativeness.
    \item 
    We propose a novel counterfactual data augmentation framework which enables reward models to generate rewards by separating content quality from response length. Training on carefully constructed response pairs that isolate one factor at a time, our method facilitates content-based preference learning during reward model training.
    \item 
    We empirically demonstrate that our approach effectively mitigates length bias and leads to more robust content-sensitive reward modeling in LLM, as evidenced by comprehensive evaluations across multiple benchmarks.
\end{itemize}

\section{Preliminaries}
\label{sec:preliminaries}

\paragraph{Length Bias in Reward Model}
Reinforcement Learning from Human Feedback (RLHF)~\citep{ouyang2022traininglanguagemodelsfollow} aligns language models with human preferences via fine-tuning on pairwise comparisons. Common approaches include Proximal Policy Optimization (PPO)~\citep{schulman2017ppo}, which uses a learned reward model, and Direct Preference Optimization (DPO)~\citep{rafailov2023direct}, which directly optimizes preferences without an explicit reward model. See~\Cref{appendix:rlhf} for further RLHF details.

Several methods have been proposed to mitigate length bias in reward modeling, primarily through architectural or training-time interventions. ODIN~\citep{chen2024odin} uses a dual-head reward model to isolate semantic and stylistic features, while RRM~\citep{liu2025rrmrobustrewardmodel} augments preference data with length perturbations to promote robustness. Although both reduce surface-level sensitivity to response length, they do \textit{not} perform controlled interventions on length itself, and may suppress stylistic variance without truly disentangling verbosity from content quality. This motivates our approach of counterfactual data augmentation, which disentangles the effect of content and length on reward in a principled manner.

\paragraph{Pearl's Causal Hierarchy}
% \label{sec:pch}
% \paragraph{Structural Causal Models}
The Pearl's Causal Hierarchy (PCH)~\citep{bareinboim20201on, pearl2018bookofwhy} organizes causal reasoning into three hierarchical levels---associational, interventional, and counterfactual---each corresponding to a distinct type of question one can pose about the world. These levels align with fundamental modes of reasoning: observing, acting, and imagining. The first level, \textit{association}, is based on statistical correlations observed in data, typically expressed as conditional probabilities such as \(P(y|x)\); for example, one may ask: “How does belief in a disease change when a particular symptom is observed?” The second level, \textit{intervention}, concerns the effects of actions or manipulations, often represented as \(P(y|do(x))\) or \(P(y_x)\), addressing questions such as: “Will the headache subside if the patient is given the drug?” The third level, \textit{counterfactual}, involves reasoning about alternate outcomes under hypothetical scenarios. Such questions are formulated as \(P(y_x|x',y')\), asking, for instance: “If the patient had taken the drug and the headache disappeared, would the headache still have persisted had they not taken the drug?” This hierarchy highlights that higher-level causal reasoning requires stronger assumptions and richer models to support counterfactual inference. See~\Cref{appendix:scm} for additional examples and formalizations.

\paragraph{Structure of Reward Model}
%\label{sec:structure-reward-model}

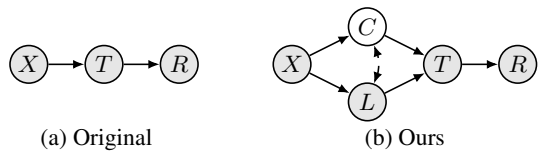
\begin{figure}[t]
\centering
\begin{tikzpicture}[
    every node/.style={font=\footnotesize, inner sep=1pt},
    graynode/.style={circle, draw=black, semithick, minimum size=5mm, fill=gray!20},
    whitenode/.style={circle, draw=black, semithick, minimum size=5mm, fill=white},
    arrow/.style={->, >={Latex[width=1.2mm,length=1.5mm]}, semithick},
    dashedarrow/.style={<->, >={Latex[width=1.2mm,length=1.5mm]}, semithick, dashed}
]

%%%%%% (a) Original (왼쪽) %%%%%%
% 노드
\node[graynode] (xa) at (0,0) {\(X\)};
\node[graynode] (ta) at (1,0) {\(T\)};
\node[graynode] (ra) at (2,0) {\(R\)};
% 엣지
\draw[arrow] (xa) -- (ta);
\draw[arrow] (ta) -- (ra);
% 캡션
\node at (0.9,-1.0) {(a) Original};

%%%%%% (b) Ours (오른쪽) %%%%%%
% 노드
\node[graynode] (xb) at (3.5,0) {\(X\)};
\node[whitenode] (cb) at ($(xb)+(1,0.5)$) {\(C\)};
\node[graynode] (lb) at ($(xb)+(1,-0.5)$) {\(L\)};
\node[graynode] (tb) at ($(xb)+(2.0,0)$) {\(T\)};
\node[graynode] (rb) at ($(tb)+(1.0,0)$) {\(R\)};
% 엣지
\draw[arrow] (xb) -- (cb);
\draw[arrow] (xb) -- (lb);
\draw[dashedarrow] (lb) to[bend right=20] (cb);
\draw[arrow] (cb) -- (tb);
\draw[arrow] (lb) -- (tb);
\draw[arrow] (tb) -- (rb);
% 캡션
\node at (5.0,-1.0) {(b) Ours};

\end{tikzpicture}
\caption{Comparison of original perspective and our perspective on reward modeling process. }
\label{fig:comparison}
\end{figure}

Common perspectives on reward modeling can be represented by the causal structure in~\Cref{fig:comparison} (a), where a response \( T \) is generated from a prompt \( X \) and then passed to a reward model which outputs a score \( R \). While the diagram shows a single response for simplicity, RLHF training is conducted on pairwise preference data, applying this structure to both responses in a comparison.\footnote{The final preference comparison between reward scores is omitted in the figure to focus on the causal pathways from the prompt \( X \) to each response \( T \).}

Crucially, however, the reward signal \( R \) is not directly supervised from the environment. Although the reward model outputs a scalar value, this value is not grounded in explicit feedback signals (as in traditional RL), but is instead inferred from binary preference labels over response pairs, either annotated by humans or provided by automated judges. As a result, the reward model is trained via comparative supervision: it learns to assign higher scores to preferred responses within triplets of the form \( (X, T_{\text{chosen}}, T_{\text{rejected}}) \). Over time, this process encourages the model to approximate a reward function that aligns with observed preferences.

\section{Causal Interpretation of Length Bias}
\label{sec:causal-interpretation}

% \subsection{Problem Setting and Formulation}
% \label{sec:problem-setting}
\paragraph{Length Bias as a Causal Problem}
% \label{sec:causal-length-bias}

% Length bias refers to the systematic tendency to assign higher scores to longer responses, even when they are no more informative than shorter ones. This phenomenon frequently arises during the reward model training process in RLHF, due to the entanglement between semantic content and response length. That is, the reward model cannot reliably distinguish whether its preference stems from semantic content or from length. Consequently, it may treat verbosity as a dominant signal for reward, leading to misaligned preferences.

% From a causal perspective, this phenomenon can be illustrated by the structure in~\Cref{fig:comparison} (b), where each response \( T \) is generated from two factors: latent semantic content \( C \) and response length \( L \). These two factors may interact, affecting each other, as responses are shaped both by what is said and how extensively it is conveyed. Prior work has shown that response length can confound human judgments of informativeness~\citep{saito2023verbositybiaspreferencelabeling, shen2023loose}, highlighting the need to decouple verbosity from semantic quality.

Length bias refers to the tendency to assign higher scores to longer responses, even when they are no more informative than shorter ones. This phenomenon frequently arises during RLHF reward model training due to the entanglement between semantic content and response length, making it difficult for the reward model to determine whether its preference stems from content or length and leading it to treat verbosity as a dominant reward signal. From a causal perspective, this can be illustrated by the structure in~\Cref{fig:comparison}(b), where each response $T$ is generated from two factors: latent semantic content $C$ and response length $L$. These two factors may interact, affecting each other, as responses are shaped both by what is said and how extensively it is conveyed. 

% Prior work has shown that response length can confound human judgments of informativeness~\citep{saito2023verbositybiaspreferencelabeling, shen2023loose}, highlighting the need to decouple verbosity from semantic quality.

\paragraph{Motivation for Counterfactuals}
% \label{sec:counterfactual-motivation}
% \paragraph{Feasibility of causal approach.}
% \label{sec:feasiblity}

% \paragraph{Geometric understanding of counterfactuals.}
% \label{sec:geometric}

\begin{figure} [t]
    \centering
    \includegraphics[width=0.65\columnwidth]{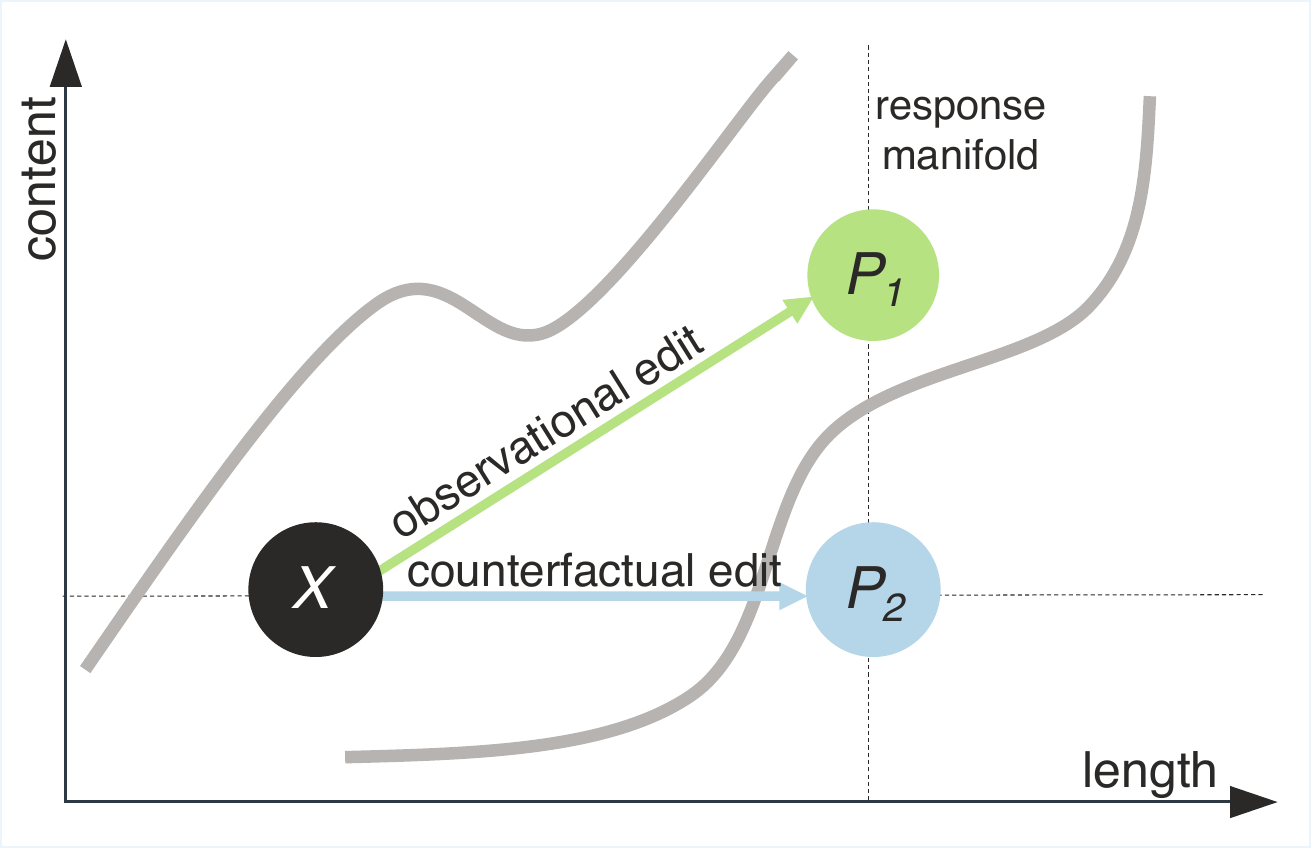}
    \caption{Response dimension over an observation edit ($P_1$) and a counterfactual ($P_2$).}
    \label{fig:response-dim}
\end{figure}

Since content and length often co-vary in natural data, conventional observational comparisons---such as randomly sampled response pairs from RLHF datasets---are insufficient to isolate the causal effect of length on reward. To address this, we introduce a \textit{counterfactual data augmentation} strategy, which generates synthetic response pairs to answer questions that natural data cannot support, such as: “What would the reward have been if the response length had been different?”

%%%%%%%% 실제로 나오는 내용은 5.approach-to-mitigate.tex지만 본문 배치를 위해 이동 
\begin{figure*}[t]
  \centering
  \includegraphics[width=0.8\textwidth]{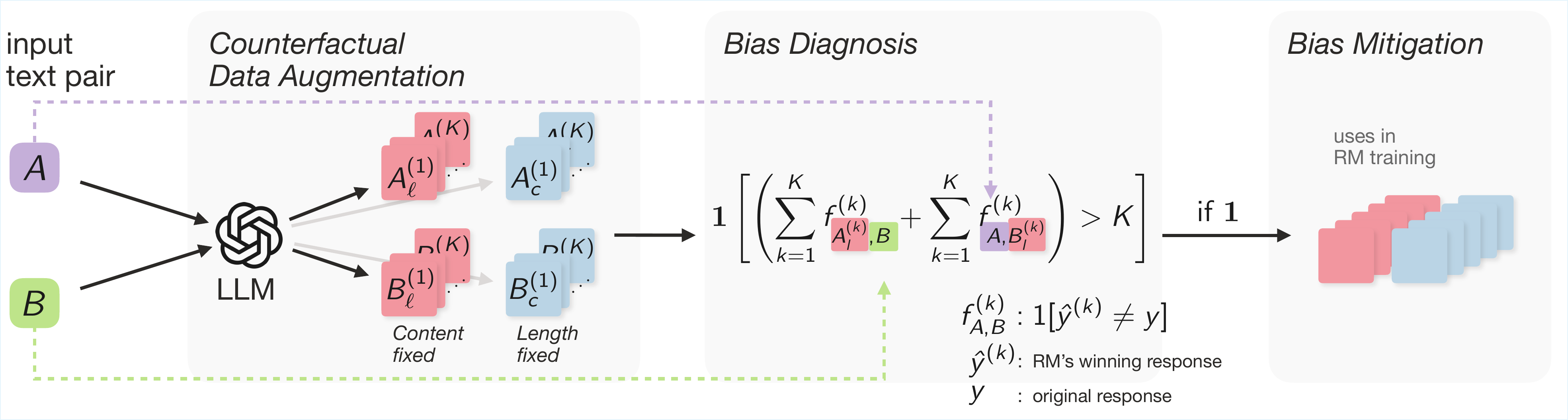}
  \caption{Overview of our method. We generate length- and content-fixed counterfactuals and use them for bias diagnosis and reward model training.}
  \label{fig:overview}
\end{figure*}
% %%%%%%%%%%%%%%%%%%%%%%%%%%%%%%%%%%%%%%%%%%%%%%%%%%%%%%%%%%%%%%%%%%%%%%%%%%%%%%%

\Cref{fig:response-dim} illustrates the need for counterfactual data augmentation geometrically. Natural responses lie on a low-dimensional manifold embedded in a space defined by content and length (~\Cref{appendix:response-dim}). Because these two attributes are often entangled in real-world data, changing one attribute (e.g., length) typically induces a shift in the other (e.g., content). For instance, an \textit{observational edit} moves along the response manifold---from \(X\) to \(P_1\)---by varying length, but often alters content implicitly. This is analogous to editing an image: adjusting hairstyle may unintentionally shift perceived gender when the underlying latent representation does not cleanly separate the two. To isolate the effect of a single factor, we must vary one while holding the other fixed. This motivates \textit{counterfactual edits}, which simulate responses \emph{off} the natural manifold (e.g., \(X \to P_2\))---something not achievable with observational data in the response manifold. 

\paragraph{Feasibility of Counterfactuals}

A natural question arises: how can one generate counterfactual data, given that counterfactuals by definition pertain to alternative outcomes that did not actually occur? In our case, such construction is feasible because the setting itself falls under the category of \textit{realizable counterfactuals}~\citep{raghavan2025counterfactual}, permitting targeted interventions (Level 2 in PCH) to approximate counterfactual outcomes (Level 3).\footnote{\citet{raghavan2025counterfactual} formally show that an L3 distribution is realizable if and only if the target variables (e.g., content \( C \) and length \( L \)) are not simultaneously subject to conflicting interventions on the same causal parent. See~\Cref{appendix:realization} for details.} We can therefore construct counterfactual responses \( T_{\tilde{c}, \ell'} \) by independently intervening on content and length: setting \( C \leftarrow \tilde{c} \) via semantic transplantation and \( L \leftarrow \ell' \) via prompt-level control. 
% This enables us to query:
% \[
% P(\text{RM}(T_{c', \ell'}) \mid C = c, L = \ell),
% \]
% i.e., “What reward would the model assign if the content and length had been \( c', \ell' \), given the original sample had \( c, \ell \)?”

When reward models are repeatedly trained on counterfactual preference pairs in which both responses have identical lengths, any residual preference signal must be attributed to semantic differences. This repeated exposure encourages the model to ground its judgments in content alone, gradually attenuating sensitivity to verbosity. In effect, this decouples response length from reward estimation without impairing semantic content of the response. As a result, the influence of length \(L\) on reward diminishes, allowing the model to better reflect content-driven distinctions. For a formal treatment of this mechanism, see~\Cref{appendix:counterfactual-formalism}.

% When reward models are repeatedly trained on counterfactual preference pairs in which both responses have identical lengths, any residual preference signal must be attributed to semantic differences. Formally, this behavior corresponds to the reward model satisfying:
% \[
% \sigma(R(T_{c, \ell}) - R(T_{c', \ell})) \gg \sigma(R(T_{c, \ell}) - R(T_{c, \ell'})),
% \]
% for all \( c \neq c', \ell \neq \ell' \), where \( \sigma(\cdot) \) denotes the sigmoid over reward difference. That is, the model becomes more sensitive to changes in content than to changes in length. This repeated exposure encourages the model to ground its judgments in content alone, gradually attenuating sensitivity to verbosity. In effect, this decouples response length from reward estimation without impairing semantic content of the response. As a result, the influence of length \(L\) on reward diminishes, allowing the model to better reflect content-driven distinctions. 

Our augmentation strategy relies on three key assumptions: (1) Each response \( T \) can be approximately decomposed into two factors: latent content \( C \) and observable length \( L \), such that their influence on the reward is fully mediated through the response; (2) It is feasible to generate alternative responses that modify one factor (e.g., verbosity) while preserving the other (e.g., semantic content), enabling controlled interventions; (3) When two responses have approximately the same length, any difference in model preference is assumed to reflect differences in semantic content quality. These assumptions allow us to treat counterfactual comparisons at fixed length as valid supervision signals for training content-aware reward models. 
% These assumptions collectively imply that learning from pairs of the form \( (T_{c, \ell}, T_{c', \ell}) \) leads to a reward function \( R \) that is approximately invariant to length:
% \[
% \frac{\partial R(T_{c, \ell})}{\partial \ell} \approx 0,
% \quad \text{while} \quad
% \frac{\partial R(T_{c, \ell})}{\partial c} \neq 0.
% \]

\subsection{Operational Definitions of Length and Content}
\label{sec:operationalization}
To enable quantifiable control required for counterfactual data augmentation, we define \textit{length} and \textit{content} in operational terms that allow consistent measurement and manipulation. For \textit{length}, we partition the empirical token distribution of responses into five quantile-based bins---Very Short, Short, Medium, Long, and Very Long---treating responses within the same bin as approximately equal in length. For \textit{content}, which is a latent property, we define it through the relational criterion of semantic equivalence. This relational definition allows us to systematically distinguish between \textit{fixed content} and \textit{varying content} pairs for counterfactual construction. \textit{Fixed Content}: Responses that convey the same meaning despite variations in tone, redundancy, or structure. \textit{Varying Content}: Responses that differ in factuality, specificity, or intent while keeping the length fixed.
\section{Length Bias Mitigation Pipeline}
\label{sec:mitigation-pipeline}
To implement our causal approach, we introduce a three-stage framework summarized in~\Cref{fig:overview}. The process consists of three main stages: (1) Counterfactual Data Augmentation: Generate augmented response variants via controlled manipulation on either response length or semantic content, keeping the other approximately constant. (2) Bias Diagnosis: Identify length-driven preference flips by applying targeted length interventions while preserving content. (3) Bias Mitigation: Retrain the reward model using curated counterfactuals that isolate semantic content from stylistic factors like verbosity.

\subsection{Counterfactual Data Augmentation Implementations}
\label{sec:counterfactual-data-augmentation}

To disentangle the effect of length on reward, we generate counterfactually augmented 
response pairs by manipulating either content or length while keeping the other factor 
approximately fixed. These augmentations produce response pairs aligned in length, 
enabling controlled comparisons under fixed-length conditions during reward learning. 
We employ two complementary augmentation strategies\footnote{Full augmentation 
procedures are provided in~\Cref{appendix:augmentation-techniques}.}:

\begin{itemize}
    \item \textit{Length-fixed augmentation}: Given a target response, we generate alternative responses that vary in factuality or informativeness by intervening on the semantic content while keeping the length approximately fixed. To achieve this, we apply transformations such as detail removal, elaboration, information substitution, or rewriting figurative expressions into literal descriptions.
    
    % \item \textit{Content-fixed augmentation}: Given a target response, we generate alternative responses that vary in verbosity by intervening on length while preserve the same underlying meaning if the original. In this case, the length is aligned to that of the reference response with which the preference label was originally determined. To preserve semantic content while varying length, we apply surface-level augmentations such as filler insertion (deletion), pleonasm (simplification), redundant sentence reusing (pruning), paraphrasing, and format changes.     
    \item \textit{Content-fixed augmentation}: Given a target response, we generate alternatives that vary in verbosity by intervening on length while preserving the original meaning. Here, the length is aligned to that of the reference response used to determine the original preference label. To preserve semantic content while varying length, we apply surface-level augmentations such as filler insertion (deletion), pleonasm (simplification), redundant sentence reusing (pruning), paraphrasing, and format changes.

\end{itemize}

After generating responses, it is essential to verify that the intended factor has been successfully manipulated. As described in~\Cref{sec:operationalization}, length is operationalized via token-level binning, enabling automatic verification through bin membership. In contrast, content is a latent property that cannot be reliably assessed using simple heuristics. To ensure fidelity, we apply automated editing strategies followed by semantic filtering using a binary classifier that checks whether semantic content is preserved. These verification steps confirm whether the intended manipulation was successful and enhance the overall quality of the augmented data.

\subsection{Diagnosing Length Bias}
\label{sec:diagnosis}

\begin{table}[t]
\footnotesize
\centering
\begin{tabular}{@{}c|ccc@{}}
\hline
\textbf{Content Quality} & \textbf{Shorter} & \textbf{Same} & \textbf{Longer} \\
\hline
Better & \cmark{} & \cmark{} & \cmark{} \\
Same   & \xmark{} & \xmark{}  & length bias \\
Worse  & \xmark{} & length bias & length bias \\
\hline
\end{tabular}
\caption{Rules for diagnosing length bias based on the winning response's content quality and relative length: the model's preference is acceptable (\cmark{}), implausible (\xmark{}), or likely indicative of length bias (length bias).}
\label{tab:win-distribution}
\end{table}

We define a diagnostic rule table (\Cref{tab:win-distribution}) that categorizes each case based on the relative content quality and length of the winning response, and determines whether the model's preference is attributable to content or likely influenced by length bias. According to this scheme, length bias is diagnosed when the model prefers a longer response with worse content, excluding ties from both diagnosis and training.

%%%%%%%%%%%%%%%%%%%%%%%%%%%%%%%%%%%%%%%%%%%%%%%%%%%%%%%%%%%
%%%%%%%%%%%%%%%%%%%%%%%%%%%%%%%%%%%%%%%%%%%%%%%%%%%%%%%%%%%
\begin{figure*} [t]
    \centering
    \includegraphics[width=0.8\textwidth]{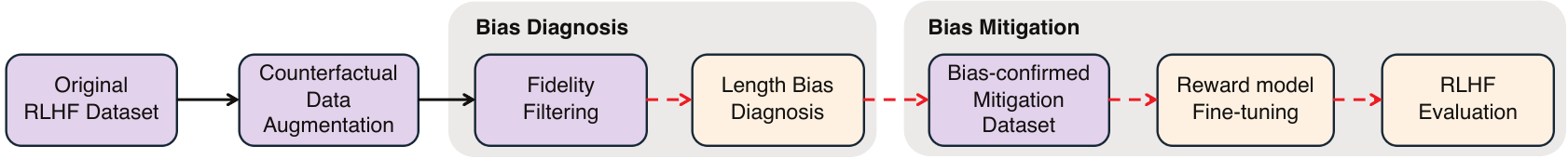}
    \caption{Experimental pipeline with edge styles indicating flow type. Purple nodes represent data transformation stages; orange nodes algorithmic processing steps. Black arrows data transformation; red dashed arrows model-based processing or state change.}
    \label{fig:pipeline-overview}
\end{figure*}
%%%%%%%%%%%%%%%%%%%%%%%%%%%%%%%%%%%%%%%%%%%%%%%%%%%%%
%%%%%%%%%%%%%%%%%%%%%%%%%%%%%%%%%%%%%%%%%%%%%%%%%%%%%

\paragraph{Preference flips.}
% To apply this diagnostic rule in practice, we measure \textit{preference flips} induced by controlled length interventions using content-fixed augmented responses. A preference flip occurs when a reward model reverses its ranking of a response pair solely due to a change in length, with content held constant. Let \( R(X, T) \) denote the reward score assigned to response \( T \) given prompt \( X \). Suppose the original model preference is \( R(X, A) > R(X, B) \). A preference flip is said to occur if, for a counterfactual variant \( A' \) or \( B' \) with identical content but altered length, the ranking reverses—e.g., \( R(X, A') < R(X, B) \) or \( R(X, A) < R(X, B') \).

% We define the \textit{flip ratio} $F$ for a pair \((A, B)\) as:
% \[F_{(A, B)} = \frac{ \# \, \text{of flipped preferences}}{\text{Total counterfactual comparisons}}\]
% Each original pair is evaluated against multiple counterfactual variants (e.g., $A$ vs $B_1'$, $A$ vs $B_2'$, etc.). We consider a pair to exhibit length bias if its flip ratio exceeds 0.5---i.e., the majority of counterfactuals trigger a reversal in preference. For example, if a model initially favors the longer response, but prefers the shorter one after a content-preserving length adjustment, we interpret this as direct evidence that the model's decision was driven by verbosity rather than semantic quality.

To scale up the diagnostic rules in \Cref{tab:win-distribution} for large-scale evaluation, we introduce a binary decision rule based on \textit{preference flips}, which are induced by controlled length interventions using content-fixed augmented responses. A preference flip occurs when the reward model reverses its ranking of a response pair solely due to a change in length, with semantic content held constant. Let \( R(X, T) \) denote the reward score assigned to response \( T \) given prompt \( X \). Suppose the original model preference is \( R(X, A) > R(X, B) \). A preference flip is said to occur if, for a counterfactual variant \( A' \) or \( B' \) with the same content but altered length, the ranking reverses—e.g., \( R(X, A') < R(X, B) \) or \( R(X, A) < R(X, B') \). When the model initially prefers the longer response but reverses its choice after a content-preserving length adjustment, we interpret this as evidence that the original preference was driven by verbosity rather than semantic quality.

For each original preference pair $(A, B)$, we generate $K$ content-fixed length variants and evaluate the model’s consistency against them (e.g., $A$ vs $B_1'$, $A$ vs $B_2'$, etc.). A pair is considered biased if the number of preference flips exceeds half of the number of counterfactuals. Formally, we apply the indicator function $\bm{1}[( \sum_{k=1}^K f^{(k)}_{A_l} + \sum_{k=1}^K f^{(k)}_{B_l} ) > K]$, where $f^{(k)}_{A_l} = \bm{1}[\hat{y}^{(k)} \neq y]$ denotes a flipped preference under the $k$-th intervention on response $A$, and $y$ is the original model preference label—i.e., the response that was favored in the original $(A, B)$ pair. This criterion detects cases where length changes alone consistently alter model decisions.

To generalize this diagnosis, we define the \textit{flip ratio} $F$: \[F_{(A, B)} = \frac{ \# \, \text{of flipped preferences}}{\text{Total counterfactual comparisons}}.\]Pairs with $F > 0.5$ are flagged as length-biased. This continuous metric enables scalable and automated diagnostics aligned with the rule-based intuition, while providing fine-grained control for downstream filtering and training.

\paragraph{Final counterfactual data used for bias mitigation.}
After identifying bias-prone examples via preference flips, we selectively use counterfactually augmented response pairs for reward model training. For response pairs that exhibit length bias---i.e., more than half of the content-fixed counterfactual pairs result in preference flips---we include the flipped examples in the training set. In addition, we incorporate the length-fixed variants of the original responses to further disentangle length from content during reward learning.

\subsection{Mitigating Length Bias} 
To illustrate how the selected counterfactually augmented data mitigates length bias, consider a response pair \((A, B)\) that has been diagnosed as length-biased through content-fixed augmentation. In the original preference data, the model favored \(A\) due to its verbosity rather than its semantic quality, meaning the original supervision signal does not reflect a true preference. To correct this, we form a new training pair by combining the content-fixed counterfactual \(A'\)---which preserves the meaning of \(A\) while matching the length of \(B\)---with the original response \(B\), forming the pair (\(A'\), \(B\)). In this counterfactual pair, where length is neutralized, we revise the 
supervision to favor \(B\), yielding a more accurate learning signal grounded in 
semantic content.

In addition to the flipped preference pairs from content-fixed augmentations, we also incorporate length-fixed augmentations to directly supervise semantic quality. For each response \(A\), we use its pre-generated length-fixed variant \(A''\), which has degraded semantic content but maintains length, and train the model to prefer \(A\) over \(A''\). This setup encourages the reward model to distinguish fine-grained semantic differences under fixed stylistic conditions, enabling learning signals that reflect content alone. For instance, if \(A''\) is semantically inferior to \(A\), and \(A\) is determined to be worse than \(B\) based on the content-controlled pair \((A', B)\), the model learns a content-grounded ranking: \(A'' < A' = A < B\).

These counterfactual augmentations ensure that the reward model learns to prioritize semantic content over superficial stylistic features such as verbosity by providing targeted supervision---via content- and length-fixed variants---that explicitly disentangles content quality from length artifacts.

\section{Experiments}
\label{sec:experiments}

We present an overview of our experimental pipeline in \Cref{fig:pipeline-overview}, which summarizes the three core stages of our method: counterfactual data augmentation, bias diagnosis, and bias mitigation through reward model fine-tuning.

\subsection{Data Augmentation}
\label{sec:augmentation}

We use the RLHF preference dataset from RLHFlow~\citep{dong2024}, consisting of 699k prompt-response pairs with pairwise preference labels from seven sources, for its large scale and diverse annotation sources. For augmentation, we selected \textbf{GPT-4o-mini}~\citep{openai2024gpt4o} due to its strong semantic fidelity and stylistic control.

To study length bias, we filtered for examples where the preferred response is longer and the two responses fall into different length bins (~\Cref{appendix:length-bins}), discarding pairs with extreme disparities ($\geq$4 bins apart). This yielded 225,358 examples, from which we randomly sampled 50,000 for augmentation. Using controlled editing strategies,\footnote{Examples and templates illustrating these augmentation strategies are provided in ~\Cref{appendix:augmentation-examples}.} we generated 474k content-fixed and 471k length-fixed response pairs, totaling approximately 945k augmented comparisons---a 19$\times$ increase over the original sample. 

To verify whether the intended factor (content or length) is correctly preserved, we fine-tuned a binary classifier based on \texttt{all-mpnet-base-v2}~\citep{song2020mpnet}.\footnote{We describe implementation details for all models—including, cross-encoders, reward models, and policy models—in~\Cref{appendix:finetuning}.} After filtering, 472k content-fixed and 466k length-fixed response pairs were retained. These filtered counterfactuals form a reliable basis for diagnosing length bias.

\subsection{Length Bias Identification and Mitigation Data Construction}
\label{sec:length-bias}
\paragraph{Length bias identification.}
To measure the presence of length bias, we remove all original preference labels from the RLHFlow dataset and re-score each prompt–response pair using a reference reward model, \texttt{OpenLLaMA-3B}~\citep{openlm2023openllama}. Then, we conducted content-fixed comparison using counterfactual responses that preserve semantic content while varying length. A \textit{flip} is recorded when a model's preference reverses due to a change in length alone. Among 49,861 pairs, 23,651 ($47.43\%$) exhibited length bias. See~\Cref{appendix:flip} for full distribution.

\paragraph{Mitigation data construction.}

\begin{table}[t]
\footnotesize
\centering
\begin{tabular}{@{}crr@{}}
\hline
\textbf{Stage} & \textbf{Content} & \textbf{Length} \\
\hline
Augmented pairs (pre-filtering)         & 474k & 471k \\
Filtered pairs   & 473k & 466k \\
Length bias pairs   & 199k & 214k \\
\hline
\end{tabular}
\caption{Summary of augmentation and filtering statistics.}
\label{tab:data-summary}
\end{table}

\begin{table*}[t]
\centering
\footnotesize
\begin{adjustbox}{max width=0.98\linewidth}
\begin{tabular}{lcccccccccccc}
\toprule
\textbf{Model} 
& \multicolumn{5}{c}{\textbf{RewardBench-1}} 
& \multicolumn{6}{c}{\textbf{RewardBench-2}} 
& \textbf{Chatbot Arena} \\
\cmidrule(lr){2-6} 
\cmidrule(lr){7-12} 
\cmidrule(lr){13-13}
& Chat & Chat Hard & Safety & Reasoning & Avg
& Factuality & PIF & Math & Safety & Focus & Avg
& LC Accuracy \\
\midrule
HRO          & \underline{0.718} & 0.485 & 0.334 & 0.420 & 0.486  & 0.364 & \textbf{0.275} & \textbf{0.350} & 0.240 & \textbf{0.238} & {0.250} & 0.249 \\
ODIN         & 0.499 & 0.487 & 0.514 & 0.485 & 0.496  & 0.301 & \underline{0.263} & 0.230 & 0.154 & 0.147 & 0.219 & 0.463 \\
CDA\_OpenLM* & 0.466 & 0.493 & \underline{0.504} & \underline{0.482} & 0.486 & \underline{0.416} & 0.257 & 0.311 & \underline{0.270} & 0.133 & \underline{0.278} & \textbf{0.508} \\
CDA\_LoRA*   & \textbf{0.732} & \underline{0.496} & 0.332 & 0.427 & \underline{0.497}  & 0.361 & 0.244 & \underline{0.336} & 0.267 & \underline{0.232} & \textbf{0.288} & 0.248 \\
CDA\_HRO*    & 0.491 & \textbf{0.510} & \textbf{0.529} & \textbf{0.495} & \textbf{0.506}  &  \textbf{0.461} & 0.197 & 0.244 & \textbf{0.281} & 0.199 & \underline{0.276} & \underline{0.493} \\
\bottomrule
\end{tabular}
\end{adjustbox}
\caption{
Accuracy of reward models across three datasets: \textbf{RewardBench-1}, \textbf{RewardBench-2}, and length-controlled (LC) accuracy from \textbf{Chatbot Arena}. (* indicates models trained with our method.)
}
\label{tab:rewardbench_transposed}
\end{table*}

For each pair classified as length-biased, we construct a mitigation dataset by combining content-fixed and length-fixed augmentations. From the content-fixed set, we retain only those that cause a preference flip, ensuring the bias is empirically observed. Then, for each confirmed flip, we include the corresponding length-fixed augmentations to reinforce content sensitivity. This yields 198,778 flipped content-fixed pairs and 213,699 aligned length-fixed augmentations. After deduplication, we obtain 412,286 unique (prompt, chosen, rejected) triplets which will be used in reward model fine-tuning. The total number of data points processed is summarized in~\Cref{tab:data-summary}.

\paragraph{Reward model finetuning.}
\label{sec:rm-finetuning}
To mitigate length bias, we fine-tune reward models on a counterfactually augmented dataset that disentangles verbosity from semantic content. We utilized two baseline models to fine-tune: \texttt{OpenLLaMA-3B} and its RLHF variant reward model, \texttt{HH-RLHF\_RM\_OpenLLaMA-3B}~\citep{diao2023lmflow,dong2023raft}.

\paragraph{Fine-tuned reward models.}
We evaluate five reward models: (1) \textbf{HRO}, the baseline reward model \texttt{HH-RLHF\_RM\_OpenLLaMA-3B}; (2) \textbf{ODIN}~\citep{chen2024odin}, a recent method which mitigates length bias via dual-head reward modeling. Although the original ODIN used Vicuna-7B, we reimplemented it on the \texttt{OpenLLaMA-3B} backbone to eliminate confounding effects from base model performance differences. This ensures that any observed differences in evaluation are attributable to methodological differences rather than disparities in model capacity or pretraining quality. (3) \textbf{CDA\_OpenLM}, a reward model obtained through fine-tuning \texttt{OpenLLaMA-3B} on our counterfactually augmented dataset; (4) \textbf{CDA\_LoRA}, a LoRA-based fine-tuning of \texttt{HRO} using our mitigation data; and (5) \textbf{CDA\_HRO}, a full fine-tuning of \texttt{HRO} on the same dataset. Comprehensive details of all evaluation experiments, including dataset specifications and repeated-run statistics, are provided in ~\Cref{appendix:alpacaeval}.

\paragraph{Evaluation of reward model length bias reduction.}
\label{sec:rm-eval}
We evaluate reward models using two complementary metrics:
\begin{itemize}
\item \textbf{RewardBench Average Score}: RewardBench~\citep{lambert2024rewardbench, malik2025rewardbench2advancingreward} provides a comprehensive evaluation of general reward model performance aligned with human preferences. We use both versions of the benchmark, with version 2 being more challenging than version 1. For consistency with our length-bias diagnosis and mitigation setup, we exclude tie cases from RewardBench-2 in our reporting.
\item \textbf{Length-controlled accuracy}: This metric directly measures the extent to which reward models rely on verbosity by evaluating whether they correctly prefer shorter responses when appropriate. Using Chatbot Arena pairwise preferences~\citep{chiang2024chatbot}, we select pairs in which the preferred response is shorter by at least two token-length bins and check whether the reward model assigns higher scores to the concise option.
\end{itemize}
Together, these metrics assess whether reducing length bias compromises general reward model performance.

\begin{figure}[t]
    \centering
    % (a) 위 이미지
    \includegraphics[width=\linewidth]{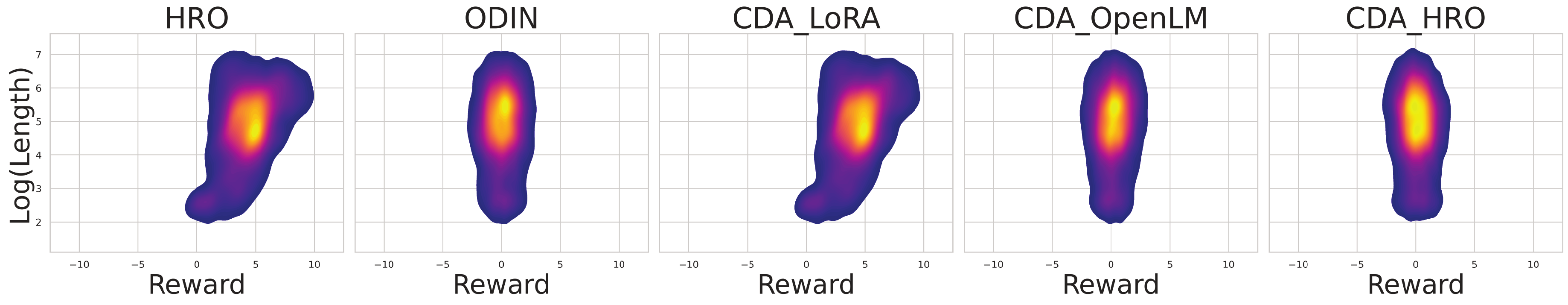}   %{\footnotesize  \par}
    % \label{fig:reward-dim-1}
    \medskip
    % (b) 아래 이미지
    \includegraphics[width=\linewidth]{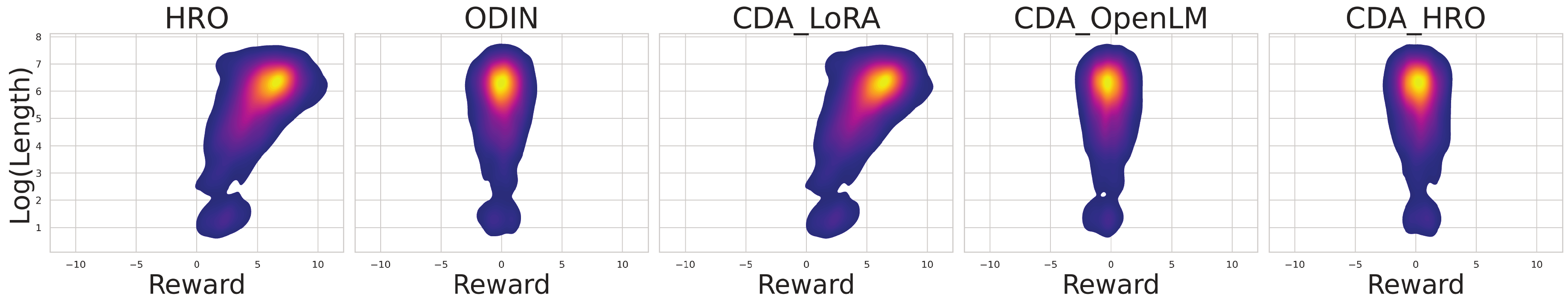}
    % {\footnotesize (b) Reward distribution across lengths on RewardBench-2 dataset. \par}
    \caption{Reward distribution across response lengths on RewardBench-1 (top) and RewardBench-2 (bottom).}
    \label{fig:reward-dim-2}

\label{fig:reward-dim}
\end{figure}
%%%%%%%%%%%%%%%%%%%%% reward dim table %%%%%%%%%%%%%%%%%%%%

% \Cref{tab:rewardbench_transposed} reports category-wise and mean accuracy on RewardBench, together with length-controlled accuracy from Chatbot Arena. On RewardBench-1, \texttt{CDA\_HRO} achieves the highest mean accuracy. Both \texttt{CDA\_HRO} and \texttt{CDA\_OpenLM} also show more balanced category-wise performance, with clear gains on bias-sensitive subsets such as \texttt{chat hard}, \texttt{safety}, and \texttt{reasoning}. These patterns indicate that our counterfactual training better captures content quality rather than surface length cues. In contrast, baseline models show more uneven category-wise behavior with similar overall scores. For RewardBench-2, \texttt{CDA\_LoRA} and \texttt{HRO} show strong peaks on certain categories, whereas \texttt{CDA\_OpenLM} and \texttt{CDA\_HRO} provide more stable performance across tasks. As RewardBench-2 consists of more challenging and less verbosity-sensitive tasks, large gains are not expected; here, our models maintain performance comparable to the baseline.

\Cref{tab:rewardbench_transposed} reports category-wise and mean accuracy on RewardBench along with length-controlled accuracy. On RewardBench-1, \texttt{CDA\_HRO} achieves the highest mean accuracy, and both \texttt{CDA\_HRO} and \texttt{CDA\_OpenLM} exhibit more balanced category-wise performance, with notable improvements on bias-sensitive subsets (\texttt{chat hard}, \texttt{safety}, \texttt{reasoning}). In contrast, baseline models display more uneven patterns with similar overall scores. For RewardBench-2, \texttt{CDA\_LoRA} and \texttt{HRO} show strong peaks in individual categories, while \texttt{CDA\_OpenLM} and \texttt{CDA\_HRO} provide more stable performance. As RewardBench-2 is more challenging and less sensitive to verbosity, large gains are not expected, and our models perform comparably to the baseline. This trend is consistent with the length-controlled evaluation, where \texttt{CDA\_OpenLM} and \texttt{CDA\_HRO} achieve substantially higher LC accuracy than the baseline, demonstrating robustness when length cannot be used as a cue. Although \texttt{CDA\_LoRA} remains competitive overall, its lower LC accuracy (24.80\%) reflects the limitations of partial fine-tuning in mitigating length bias. Across both RewardBench versions, our CDA-based models maintain competitive overall accuracy with more stable category-wise behavior, while substantially improving length-controlled accuracy. 

Taken together, these results show that our approach mitigates length bias without compromising general reward model performance, overcoming the trade-off commonly observed in baseline methods.

\paragraph{Reward distribution across length.}
To examine how reward models handle verbosity, we visualize reward–length distributions for RewardBench-1 and the more challenging RewardBench-2 (\Cref{fig:reward-dim}). In both datasets, \texttt{HRO} shows a strong positive correlation between response length and reward, revealing substantial length bias. In contrast, our counterfactually fine-tuned models (\texttt{CDA\_HRO}, \texttt{CDA\_OpenLM}) and ODIN yield more vertically aligned distributions, indicating reduced sensitivity to verbosity. The gap becomes even clearer on the harder RewardBench-2, where the baseline’s bias intensifies while our models remain robust.

\paragraph{Policy model finetuning.}
\label{sec:ppo}
To evaluate the downstream impact of length bias mitigation, we fine-tune a policy via Supervised Fine-Tuning (SFT) followed by PPO~\citep{schulman2017ppo}, using reward models with and without mitigation. This allows us to assess how improvements in reward modeling affect final policy behavior.

We evaluate six RLHF policy models, all initialized from the same SFT model based on \texttt{OpenLLaMA-3B}\footnote{Note that we exclude \texttt{CDA\_LoRA} from PPO training, as parameter-efficient fine-tuning under frozen reward heads showed limited effectiveness in mitigating bias, despite its overall competitive performance (\Cref{tab:rewardbench_transposed}).}:
(1) \textbf{OpenLM}, the unaligned base OpenLLaMA-3B model;
(2) \textbf{SFT}, trained on supervised instruction-following data;
(3) \textbf{PPO\_HRO}, fine-tuned with the baseline reward model \texttt{HRO};
(4) \textbf{ODIN}, trained using the ODIN reward model reimplemented on OpenLLaMA-3B;
(5) \textbf{PPO\_CDA\_OpenLM}, trained with our counterfactual reward model \texttt{CDA\_OpenLM}; and
(6) \textbf{PPO\_CDA\_HRO}, trained with \texttt{CDA\_HRO}.
This setup enables controlled comparison of reward strategies and the effectiveness of counterfactual data.

\paragraph{Final RLHF performance.}
\label{sec:final-rlhf}
We follow the AlpacaEval protocol~\citep{dubois2025lengthcontrolledalpacaevalsimpleway}, where models are evaluated against \texttt{LLaMA-2-7B-chat-hf}~\citep{touvron2023llama}. Since our base policy model is OpenLLaMA-3B, which is derived from LLaMA-1, we select LLaMA-2-7B-chat as a reference—upgrading both the version (LLaMA-1 to LLaMA-2) and model size (3B to 7B). This choice ensures a stronger and more up-to-date judge model for comparison. Additional evaluations against more recent baselines are reported in~\Cref{appendix:recent-baselines}.

\begin{table}[t]
\footnotesize
\centering
\setlength{\tabcolsep}{2pt} % 열 간격 조절
\begin{adjustbox}{width=\columnwidth}
\begin{tabular}{@{}lccc@{}}
\hline
\textbf{Model} & \textbf{Length Controlled Winrate} & \textbf{Winrate} & \textbf{Avg. length} \\
\hline
OpenLM     & 8.47 & 9.94 & 1385     \\
SFT     & 16.97 & 25.71 & 2061     \\
PPO\_HRO   & 18.97 & 28.45 & 2048 \\
ODIN     & 12.19 & 11.34 & 1026     \\
PPO\_CDA\_OpenLM\textsuperscript{*}    & \underline{36.06} & \underline{30.69} & 1072     \\
PPO\_CDA\_HRO\textsuperscript{*}     & \textbf{37.18} & \textbf{32.55} & 1118 \\
\hline
\end{tabular}
\end{adjustbox}
% }
\caption{Length controlled winrate of RLHF models on AlpacaEval. (* for ours.)}
\label{tab:alpaca_results}
\end{table}

% \Cref{tab:alpaca_results} shows that our method, \texttt{PPO\_CDA\_HRO}, achieves the highest length-controlled winrate (37.18\%) on AlpacaEval, more than doubling the performance of baselines like \texttt{PPO\_HRO} (18.97\%) and \texttt{SFT} (16.97\%). While they exhibit relatively high overall winrates, their lower performance under length control indicates reliance on response length. This demonstrates that counterfactual data augmentation effectively mitigates length bias, encouraging concise yet high-quality responses. \texttt{ODIN} reduces average response length but sacrifices winrate, showing a tradition trade-off between length and performance. In contrast, our models achieve strong performance without sacrificing quality for brevity, suggesting that counterfactual data augmentation aligns model preferences with content quality rather than surface features.

\Cref{tab:alpaca_results} shows that our RLHF-trained model, \texttt{PPO\_CDA\_HRO}, achieves the highest length-controlled winrate (37.18\%), more than doubling that of \texttt{PPO\_HRO} (18.97\%) and \texttt{SFT} (16.97\%). While baseline models achieve reasonable overall winrates, their sharp drop under length control indicates reliance on verbosity. In contrast, our CDA-trained policies produce shorter yet higher-quality responses, demonstrating that gains at the reward-model level carry over to downstream RLHF behavior. Unlike ODIN, which reduces length at the cost of winrate, our method improves conciseness without degrading performance.

\begin{figure} [t]
    \centering
    \includegraphics[width=\linewidth]{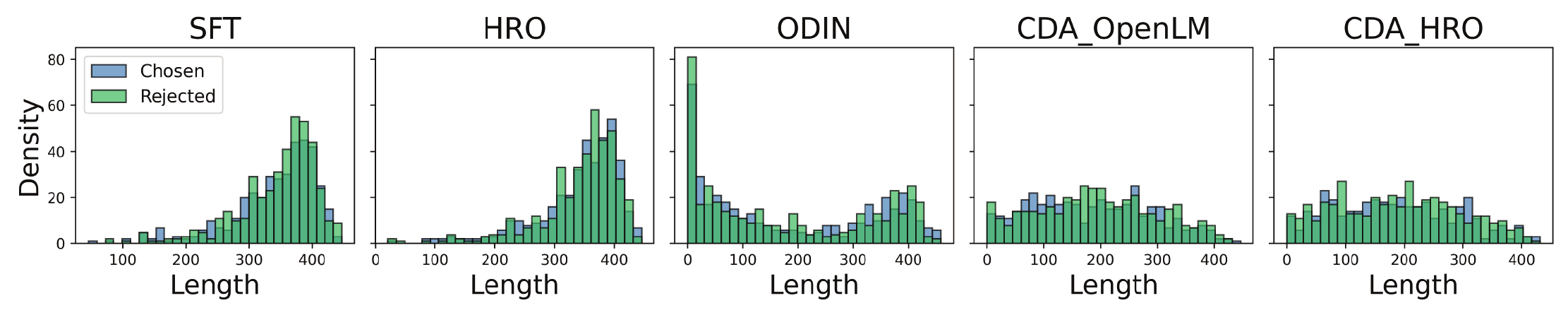}
    \caption{Length distribution of model outputs. OpenLM is excluded from this comparison as it is an untrained baseline model without any supervised fine-tuning.}
    \label{fig:length-histogram}
\end{figure}

To identify factors driving win rate differences in AlpacaEval, we analyzed response-length distributions. As shown in Figure~\ref{fig:length-histogram}, traditional reward models like \texttt{SFT} and \texttt{PPO\_HRO}, tend to favor longer responses, with a noticeable skew toward higher word counts. In contrast, models trained with our counterfactual data augmentation approach (\texttt{PPO\_CDA\_OpenLM} and \texttt{PPO\_CDA\_HRO}) show a more balanced distribution, with shorter, content-rich responses being selected more frequently. This indicates that our improved performance stems from the ability to generate more concise, high-quality outputs rather than simply producing longer responses.

\paragraph{Summary.}
Taken together, our results show that counterfactual reward-model fine-tuning improves robustness to length bias without sacrificing overall performance. Across reward-level benchmarks and downstream RLHF evaluation, CDA-based models avoid verbosity-driven preferences and enable policies to generate more concise, content-faithful responses, outperforming prior approaches.

\section{Conclusion}
\label{sec:conclusion}

We presented a causal framework for mitigating length bias in reward models trained through reinforcement learning from human feedback (RLHF). Our method uses counterfactual data augmentation to disentangle the effect of semantic content from verbosity, enabling reward models to better reflect human preferences grounded in meaning rather than surface features. By generating content-fixed and length-fixed response pairs, we allow the reward model to learn preferences that are robust to stylistic confounds and focused on underlying semantic quality.
Empirical evaluations demonstrate that our approach substantially reduces length-driven preference errors while maintaining or improving alignment performance across standard RLHF benchmarks. In particular, models trained with our counterfactually augmented data consistently prefer more concise yet informative responses and show greater robustness to stylistic variations. Moreover, downstream policy models optimized with these improved reward signals generate responses that are not only shorter on average but also rated higher in informativeness and relevance. These findings underscore the utility of causality in enhancing the fidelity and interpretability of reward models. While our method assumes a clean separation between content and length, it opens the door for future extensions that address additional confounding factors—such as tone, coherence, or factuality—through appropriate expansions of the underlying causal graph.


\section*{Acknowledgments}
We thank anonymous reviewers for constructive comments to improve the manuscript. 
This work was supported
by NRF (RS-2023-00211904/50\%, RS-2023-00222663/50\%) grant funded by
the Korean government.

\bibliography{aaai2026}

% \input{AAAI2026/ReproducibilityChecklist}

\appendix
\setcounter{secnumdepth}{2}
\onecolumn
\section*{Code and Data Availability}
All code, counterfactual augmentation scripts, and evaluation pipelines used in this work are publicly available at: \texttt{https://github.com/hazelkimm/causalRLHF}.

\section{Reinforcement Learning from Human Feedback (RLHF)}
\label{appendix:rlhf}
RLHF aligns language model outputs with human preferences by optimizing over learned reward signals derived from human-provided comparisons~\citep{christiano2017deep, ouyang2022traininglanguagemodelsfollow}. The core components involve: (i) training a reward model to predict pairwise human preferences, and (ii) optimizing the language model policy using reinforcement learning with the reward model as a proxy for human preference.

\paragraph{Reward Model Formulation.}
The reward model assigns a scalar score \( R(X, T) \in \mathbb{R} \) to a prompt–response pair, where \( X \) is the input prompt and \( T \) is the model-generated response. Human preference annotations are typically collected in the form of pairwise comparisons between two candidate responses \( T_1 \) and \( T_2 \), modeled using a Bradley–Terry style preference likelihood~\citep{bradley1952rank}:
\[
P(T_1 \succ T_2 \mid X) = \sigma(R(X, T_1) - R(X, T_2)),
\]
where \( \sigma(\cdot) \) denotes the sigmoid function.

Given a dataset of such comparisons, the reward model is trained using a margin-based ranking loss~\citep{ouyang2022traininglanguagemodelsfollow}:
\[
\mathcal{L}_{\text{RM}} = \max\left(0, m - R(X, T_{\text{chosen}}) + R(X, T_{\text{rejected}})\right),
\]
where \( T_{\text{chosen}} \succ T_{\text{rejected}} \), and \( m > 0 \) is a margin hyperparameter.

\paragraph{Policy Optimization via PPO.}
While several methods have been proposed for policy optimization in the RLHF framework---including direct approaches such as Direct Preference Optimization (DPO)~\citep{rafailov2023direct} and implicit alignment methods---this work adopts the standard Proximal Policy Optimization (PPO)~\citep{schulman2017ppo}, a widely used actor-critic algorithm known for its training stability and sample efficiency.

After training the reward model, we fine-tune the language model policy \(\pi_\theta\) using PPO, which optimizes the following clipped objective:
\[
\mathcal{L}_{\text{PPO}} = -\hat{\mathbb{E}}_t \left[ \min\left(r_t(\theta) \hat{A}_t, \text{clip}(r_t(\theta), 1 - \epsilon, 1 + \epsilon)\hat{A}_t\right) \right],
\]
where \( r_t(\theta) = \frac{\pi_\theta(a_t \mid s_t)}{\pi_{\theta_{\text{old}}}(a_t \mid s_t)} \) is the probability ratio between the current and previous policies, and \( \hat{A}_t \) is an estimate of the advantage function. The reward used to compute \( \hat{A}_t \) is provided by the learned reward model \( R(X, T) \), which assigns scalar scores to prompt–response pairs.

\paragraph{Disentangling Length and Content.}
Recent works have identified that reward models trained on human preferences often exhibit a \textit{length bias}—a systematic tendency to assign higher rewards to longer responses regardless of content quality~\citep{chen2024odin, liu2025rrmrobustrewardmodel}. In this context, our objective is to learn a reward function \( R(X, T) \) that is sensitive to the underlying semantic content \(C\) but invariant to stylistic or superficial properties such as response length \(L\). Formally, we aim to satisfy:
\[
\frac{\partial R(X, T)}{\partial L} \approx 0 \quad \text{while} \quad \frac{\partial R(X, T)}{\partial C} \neq 0.
\]
To achieve this, we introduce a counterfactual data augmentation approach that generates matched response pairs differing in length but holding content approximately constant, enabling disentangled estimation of reward signals and reducing the reliance on length-correlated artifacts in human feedback.

\section{Counterfactual Reasoning in Structural Causal Models}
\label{appendix:scm}

This appendix formalizes the structural foundations of causal reasoning used throughout the paper. We begin by introducing \textbf{Structural Causal Models (SCMs)}, which provide the mathematical basis for interventions and counterfactual queries. We then describe \textbf{Pearl’s Causal Hierarchy (PCH)}, which classifies types of causal questions into a three-level hierarchy. Finally, we relate these ideas to reward modeling in RLHF and explain why counterfactual reasoning is required to mitigate length bias.

\subsection{Structural Causal Models}
\label{appendix:scm-def}

A \textit{structural causal model} (SCM)~\citep{pearl1995causal, pearl2000causality} is a formal framework for modeling causal relationships among variables in a system. Unlike purely statistical models that capture correlations, SCMs explicitly encode how variables influence each other through a system of structural equations. This makes them suitable for answering causal queries, including those involving interventions and counterfactuals. Formally, a structural model \(\mathcal{M} \) is defined as a tuple \( \mathcal{M} = \langle \mathbf{U}, \mathbf{V}, \mathcal{F}, P(\mathbf{U}) \rangle \), where:
\begin{itemize}
    \item $\mathbf{U}$ denotes a set of \textit{exogenous variables}, whose values are determined by factors external to the model;
    \item $\mathbf{V} = \{V_1, V_2, \dots, V_n\}$ is a set of \textit{endogenous variables}, whose values are determined by variables in $\mathbf{U} \cup \mathbf{V}$;
    \item $\mathcal{F} = \{f_1, f_2, \dots, f_n\}$ is a set of structural assignments such that each $f_i$ maps from the domains of a subset of exogenous variables $\mathbf{U}_i \subseteq \mathbf{U}$ and a set of parent endogenous variables $\mathbf{Pa}_i \subseteq \mathbf{V} \setminus \{V_i\}$ to $V_i$. Each structural equation is written as $v_i \leftarrow f_i(\mathbf{pa}_i, \mathbf{u}_i)$;
    \item $P(\mathbf{U})$ is a probability distribution over the exogenous variables.
\end{itemize}

\subsection{Formalization of Pearl's Causal Hierarchy}
\label{appendix:pch-def}

\begin{figure}[t]
\centering
\begin{tikzpicture}[
    every node/.style={font=\footnotesize},
    box/.style={draw, rounded corners, minimum width=3.2cm, minimum height=1cm, align=center},
    level/.style={draw, rounded corners, minimum width=4.3cm, minimum height=1cm, align=center, fill=gray!10},
    arrow/.style={->, semithick, >=Latex}
]

% SCM box on the left
\node[box] (scm) at (0, 0) {
    \textbf{SCM (Unobserved Nature)}\\[2pt]
    $X \leftarrow f_X(U_x)$\\
    $Y \leftarrow f_Y(X, U_y)$\\
    \\
    $P(U_x, U_y)$
};

% Level 1 to the right
\node[level] (assoc) at (5.0, 1.5) {
    \textbf{Level 1: Associational}\\[2pt]
    $P(Y \mid X)$
};
% Level 2
\node[level] (interv) at (5.0, 0.0) {
    \textbf{Level 2: Interventional}\\[2pt]
    $P(Y \mid \text{do}(X))$
};
% Level 3
\node[level] (cf) at (5.0, -1.5) {
    \textbf{Level 3: Counterfactual}\\[2pt]
    $P(Y_x \mid X = x', Y = y')$
};

% Arrows from SCM to each level
\draw[arrow] (scm.east) -- ++(0.3,0) |- (assoc.west);
\draw[arrow] (scm.east) -- ++(0.3,0) -- (interv.west);
\draw[arrow] (scm.east) -- ++(0.3,0) |- (cf.west);

\end{tikzpicture}
\caption{Pearl’s Causal Hierarchy as derived from a Structural Causal Model (SCM). Each level corresponds to a distinct class of causal queries: observed associations, hypothetical interventions, and imagined counterfactuals.}
\label{fig:pch-general}
\end{figure}
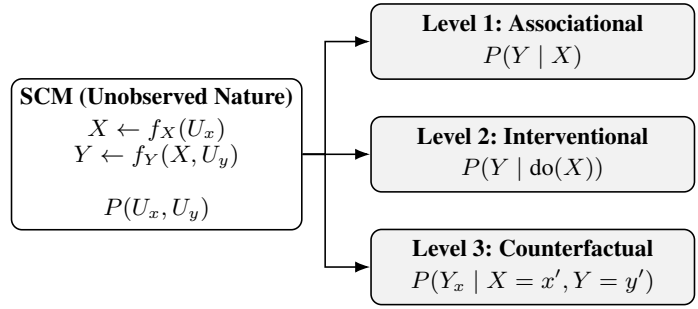

Modern causal inference begins with a key philosophical insight: that observed data and the latent mechanisms that generate them are fundamentally distinct. While data encode statistical regularities—such as correlations  between variables—they alone are insufficient to uncover the causal processes that underlie those patterns. This distinction is formally captured by the framework of \textit{Structural Causal Models (SCMs)}~\citep{pearl1995causal, pearl2000causality}, which represent data-generating processes as deterministic functions of latent variables, subject to probabilistic noise. SCMs not only describe observable associations, but also support reasoning about external interventions and counterfactual outcomes.

As illustrated in Figure~\ref{fig:pch-general}, SCMs induce what is known as the \textit{Pearl Causal Hierarchy (PCH)}~\citep{pearl2018bookofwhy}, which organizes causal queries into three distinct levels of expressiveness:
\begin{itemize}
    \item \textbf{Level 1 (Associational)}: Queries of the form \( P(\mathbf{Y} \mid \mathbf{X}) \), which capture statistical dependencies directly estimable from observed data.
    \item \textbf{Level 2 (Interventional)}: Queries like \( P(\mathbf{Y} \mid \text{do}(\mathbf{x})) \), which describe the effect of manipulating a variable \( \mathbf{x} \) while holding others fixed.
    \item \textbf{Level 3 (Counterfactual)}: Queries such as \( P(\mathbf{Y_x} \mid \mathbf{X} = \mathbf{x'}, \mathbf{Y} = \mathbf{y'}) \), which imagine alternate outcomes under hypothetical interventions.
\end{itemize}

Each level requires progressively stronger assumptions and modeling capabilities. Level 1 assumes only access to data; Level 2 additionally requires a causal graph or structural equations; Level 3 further relies on assumptions about latent variables and their role in determining individual-level outcomes.

\paragraph{Level 1: Association ("Seeing")}
\label{appendix:pch-layer1}

Layer 1 of the Pearl Causal Hierarchy deals with purely observational quantities, i.e., statistical associations among variables as they appear in the data. These are expressions of the form \( P(\mathbf{Y} \mid \mathbf{X}) \), which reflect the likelihood of an outcome \( Y \) given an observed condition \( X \). In the language of SCMs \( \mathcal{M} = \langle \mathbf{U}, \mathbf{V}, \mathcal{F}, P(\mathbf{U}) \rangle \), Layer 1 valuations are defined as:
\[P^\mathcal{M}(\mathbf{y}) = \sum_{\mathbf{u} \,|\, \mathbf{Y}(\mathbf{u}) = \mathbf{y}} P(\mathbf{u}),\]
where \( \mathbf{Y}(\mathbf{u}) \) is the solution of the structural equations under a given assignment \( u \in \mathbf{U} \).

For example, Suppose two exogenous variables \( U_1, U_2 \sim \text{Uniform}\{1, \dots, 6\} \) represent dice rolls, and the endogenous variables are defined as: \[X = U_1 + U_2,\quad Y = U_1 - U_2.\]
Given that \( P(U_1 = 1, U_2 = 1) = \frac{1}{36} \), and that this is the only joint assignment of \( U_1 \) and \( U_2 \) capable of generating the observed outcome \( \langle X = 2, Y = 0 \rangle \), it follows directly that the probability of this observation is also \( P(X = 2, Y = 0) = \frac{1}{36} \)~\citep{pearl2000causality}. More generally, Layer 1 corresponds to passive learning and inference~\citep{bareinboim20201on}. It allows answering questions like: “How likely is it to see \( Y = y \) when \( X = x \) is observed?”, or, “What is the distribution of recovery rates among patients who happened to receive the drug?”

\paragraph{Level 2: Interventional Reasoning ("Doing")}
\label{appendix:pch-layer2}

Layer 2 introduces the concept of external manipulation through the \textit{do}-operator. Queries in this layer take the form \( P(\mathbf{Y} \mid \text{do}(\mathbf{X} = \mathbf{x})) \), or \(P(\mathbf{Y_x})\), which answers: “What happens to \( Y \) if we force \( X \) to be \( x \)?” Formally, intervening on \( X \) modifies the SCM by replacing the original structural assignment \(\mathcal{F}\) with a constant:
\[\mathcal{M}_\mathbf{x} = \langle \mathbf{U}, \mathbf{V}, \mathcal{F}_\mathbf{x}, P(\mathbf{U}) \rangle,\quad \text{where}\quad \mathcal{F}_\mathbf{x} = \{ f_i : V_i \notin \mathbf{X} \} \cup \{ \mathbf{X} \leftarrow {\mathbf{x}} \}.\]
The corresponding interventional distribution is:
\[P^\mathcal{M}(\mathbf{y}_\mathbf{x}) = \sum_{\mathbf{u} \,|\, \mathbf{Y_x}(\mathbf{u}) = \mathbf{y}} P(\mathbf{u}).\]

This process disconnects \( X \) from its original causal parents and thus isolates its direct causal effect on \( Y \)~\citep{pearl1995causal, pearl2000causality}. Unlike Layer 1, Layer 2 allows for modeling "what-if" effects based on actions. For instance, if a patient receives a drug \((X = 1)\)  due to a physician's decision (not because of the patient’s prior symptoms), then \( P(Y \mid \text{do}(X = 1)) \) reflects the expected outcome of the treatment itself, not merely associations found in the data. This layer is essential for tasks such as policy evaluation (e.g., “What would happen if all patients received drug A?”), experimental design (e.g., A/B testing), and causal effect estimation (e.g., Average Treatment Effect).

\paragraph{Level 3: Counterfactual Reasoning ("Imagining")}
\label{appendix:pch-layer3}

Layer 3 of the Pearl Causal Hierarchy captures the most expressive form of causal reasoning: counterfactual inference. This level enables reasoning about alternate realities—what \emph{would} have happened under different conditions—by evaluating expressions of the form:
\[P(\mathbf{Y_{x}} \mid \mathbf{X = x'}, \mathbf{Y = y'}),\]
which quantifies: “Given that we observed \( X = x' \) and \( Y = y' \), what is the probability that \( Y \) would have taken the value \( y \) had \( X \) instead been set to \( x \)?”

Formally, an SCM \( \mathcal{M} = \langle \mathbf{U}, \mathbf{V}, \mathcal{F}, P(\mathbf{U}) \rangle \) defines joint distributions over counterfactuals via:
\[P^\mathcal{M}(\mathbf{y}_\mathbf{x},\dots, \mathbf{z_w}) = \sum_{\mathbf{u} \,|\, \mathbf{Y_x(u)} = \mathbf{y}, \dots, \mathbf{Z_w(u)} = \mathbf{z}} P(\mathbf{u}),\]
where each subscript (e.g., \( \mathbf{x} \), \( \mathbf{w} \)) denotes a distinct hypothetical intervention. The evaluation proceeds in three steps \citep{pearl2000causality, bareinboim20201on}:
\begin{enumerate}
    \item For each intervention (e.g., \( \text{do}(\mathbf{X} = \mathbf{x}) \)), replace the corresponding structural equations \( \mathcal{F}_\mathbf{X} \) with constant assignments, yielding modified functions \( \mathcal{F}_\mathbf{x} \) and creating submodels \( \mathcal{M}_\mathbf{x} \).
    \item For each background variable assignment \( \mathbf{u} \in \mathbf{U} \), evaluate the modified mechanisms \( \mathcal{F}_\mathbf{x}, \dots \, \mathcal{F}_\mathbf{w} \) in a valid causal order, solving for counterfactual outcomes \( \mathbf{Y_x}(\mathbf{u}), \dots, \mathbf{Y_w}(\mathbf{u})\).
    \item Accumulate the prior probability mass \( P(\mathbf{U} = \mathbf{u}) \) for each instance \( u \) consistent with the specified counterfactual conditions.
\end{enumerate}

For example, suppose a patient did not receive a treatment and subsequently died, i.e., \( (X=0, Y=0) \). A counterfactual query of the form
$P(Y_{X=1} = 1 \mid X = 0, Y = 0)$
asks: “What is the probability that the patient would have survived had they been given the treatment?” This query spans two distinct worlds: the factual world (\( X=0, Y=0 \)) and the counterfactual world under \( X=1 \), in which we evaluate whether \( Y=1 \) would have occurred. Such reasoning is essential in domains like legal attribution (e.g., tort law), retrospective diagnosis, and algorithmic fairness (e.g., evaluating whether outcomes would differ for counterfactual identities such as gender or race).

Unlike associative and interventional queries, counterfactuals cannot be resolved from observational or experimental data alone. They require full specification of the causal mechanisms encoded in the SCM, including assumptions about unobserved confounding and functional dependencies~\citep{pearl2009causality}.

\section{Response Manifold}
\label{appendix:response-dim}

% \begin{figure}[t]
% \centering
% \begin{tikzpicture}
% \begin{axis}[
% view={120}{30},
% xlabel={Content ($\tilde{C}$)},
% ylabel={Length ($L$)},
% zlabel={Reward},
% colormap/viridis,
% grid=major,
% domain=0:1,
% y domain=0:1,
% samples=30,
% samples y=30,
% zmin=0, zmax=1,
% hide axis=false,
% width=0.6\columnwidth,
% height=5cm,
% ]

% % Normalized Reward surface
% \addplot3[
% surf,
% shader=interp,
% ]
% {0.5 * (sin(deg(pi*x)) + y^2)};

% % Points
% \addplot3+[only marks, mark=*, mark size=2pt, color=blue] coordinates {(0.2, 0.2, 0.01)};
% \node at (axis cs:0.2, 0.2, 0.1) [anchor=east, color=blue, font=\bfseries\large] {$X$ ($\tilde{c}$, $l$, $z_0$)};

% % P1 ON surface
% \addplot3+[only marks, mark=*, mark size=2pt, color=red] coordinates {(0.2, 0.6, 0.4)};
% \node at (axis cs:0.1, 0.3, 0.4) [anchor=west, color=red, font=\bfseries\large] {P1 ($c'$, $l'$, $z_1$)};

% % P2 BELOW surface
% \addplot3+[only marks, mark=*, mark size=2pt, color=green!50!black] coordinates {(0.3, 0.7, 0.1)};
% \node at (axis cs:0.3, 0.7, 0.13) [anchor=north, yshift=-4pt, color=black, font=\bfseries\large] {P2 ($\tilde{c}$, $l'$, $z_2$)};

% % Arrows
% \addplot3[->, thick, color=red] coordinates {
% (0.2, 0.2, 0.01)
% (0.2, 0.6, 0.4)
% };

% \addplot3[->, thick, dashed, color=green!50!black] coordinates {
% (0.2, 0.2, 0.01)
% (0.3, 0.7, 0.1)
% };

% \end{axis}
% \end{tikzpicture}
% \caption{
% Reward over content ($\tilde{C}$) and length ($L$). P1 (red) is an observational edit, and P2 (green) is a counterfactual.
% }
% \label{fig:3d-reward}
% \end{figure}

In the context of language model reward modeling, we use the term \textit{response manifold} to refer to a low-dimensional surface embedded in a space defined by latent semantic content \( C \in \mathcal{C} \) and stylistic attributes such as length \( L \in \mathcal{L} \). This idea builds on foundational insights from representation learning, where high-dimensional data (e.g., text) often concentrates around structured, smooth manifolds~\citep{bengio2013representation, higgins2017beta, tishby2000information}. Let \( \mathcal{M} \subset \mathbb{R}^2 \) denote the empirical approximation to this response manifold—namely, the support of the distribution \( \mathcal{D}(C, L) \), where each point corresponds to a model-generated response characterized by a content–length pair \( (c, \ell) \). For analytical clarity, we treat \(C\) and \(L\) as coordinates in a continuous latent vector space, so that each response is represented as a point \((c, \ell) \in \mathbb{R}^2\) and expectations and distances are taken in this latent space rather than over raw text.

% Let \( \mathcal{M} \subset \mathbb{R}^2 \) denote the support of the empirical distribution \( \mathcal{D}(C, L) \), where each point corresponds to a model-generated response characterized by a content-length pair \( (c, \ell) \).

For autoregressive language models, responses are sampled from conditional distributions shaped by the input prompt and learned parameters, which may have been fine-tuned using reinforcement learning (RL) objectives in some cases. While the full space of text is vast and high-dimensional, model-generated outputs tend to form trajectories with correlated changes in textual attributes~\citep{hejna2024, ganguli_2022}---what we refer to as the \textit{response manifold}. Along this manifold, natural variation in one attribute (e.g., length) typically co-occurs with variation in others (e.g., semantic richness or syntactic complexity).

From a causal perspective, responses sampled from this manifold \( \mathcal{M} \) represent \textit{observational data}: they reflect joint variation of multiple factors under natural generation dynamics. Therefore, any edit along the manifold—such as making a response longer—implicitly affects entangled latent features. This makes it difficult to isolate the causal effect of individual attributes like verbosity on reward assignment~\citep{pearl2009causality, schölkopf2021}.

\paragraph{Observational edits as manifold projections.}
Given a response \( (c, \ell) \in \mathcal{M} \), we define an observational edit to new length \( \ell' \) as:

\[
(c', \ell')_{\text{obs}} = \arg\min_{(c', \ell') \in \mathcal{M}} \left\Vert \mathbb{E}_{(c, \ell) \sim \mathcal{D}} \left[ (c, \ell) \mid L = \ell' \right] - (c', \ell') \right\Vert,
\]

where the updated response moves along the data manifold and reflects the conditional distribution over content given length. Such edits tend to entangle length and content, reinforcing spurious associations in reward learning.
To isolate such causal effects, we require response variants where only one factor is perturbed. We formalize this idea as \textit{counterfactual-style edits} below.

\paragraph{Counterfactual-style edits as controlled perturbations.}
\[
(\tilde{c}, \ell')_{\text{cf}} = \text{synthetic response with fixed content and modified } L = \ell'.
\]

Here, \( \tilde{c} \) denotes a style-invariant core meaning of the original response, which should remain unaffected by variations in length. These examples may lie off-manifold (i.e., outside \( \mathcal{M} \)) and are unlikely under the natural data distribution. While not sampled from the true data distribution, such edits enable the reward model to learn more disentangled, content-sensitive reward signals. These counterfactual-style edits can be approximated through synthetic rewrites, disentangled generation, or latent-structure-guided data augmentation~\citep{locatello2019, cai2025}.

In our framework, we exploit this idea by constructing training pairs that approximate such counterfactual perturbations—e.g., responses with similar content but different lengths, or vice versa. This provides the reward model with examples that simulate targeted interventions, allowing it to learn disentangled reward signals that better reflect human preferences for substance over verbosity. This controlled augmentation strategy enables the reward model to attend to semantically meaningful differences, reducing reliance on stylistic proxies such as verbosity.

\section{Counterfactual Realizability}
\label{appendix:realization}

Under Pearl’s Causal Hierarchy (PCH)~\citep{pearl2018bookofwhy}, Layer 3 ($\mathcal{L}_3$) queries—counterfactuals such as \( P(Y_x \mid X = x', Y = y') \)—cannot be answered from observational or interventional data alone. Traditionally, such quantities are regarded as non-realizable, as they involve conflicting realities (i.e., an individual observed under \( X = x \), but imagined under \( \text{do}(X = x') \)). However, recent work by~\citet{raghavan2025counterfactual} provides formal conditions under which $\mathcal{L}_3$-distributions are \textit{realizable} through physical experimentation. This section formalizes realizability and shows that our experimental setup for length bias mitigation satisfies these conditions. The following definitions, assumptions, and corollaries are based on the work of~\citet{raghavan2025counterfactual}.

\paragraph{Definition of realizability.}
Let \(\mathcal{G} \) be a causal diagram, and \( \mathbb{A} \) a set of physical actions (e.g., standard interventions, counterfactual randomizations) that an agent can perform in the environment. Let \( \mathcal{M} \in \mathcal{M}(\mathcal{G}) \) denote a structural causal model consistent with \( \mathcal{G} \). Before introducing realizability, we recall the notion of an i.i.d.\ sample from an \(\mathcal{L}_3\)-distribution:

\begin{quote}
\textbf{Definition (I.i.d.\ sample, Def. 3.3):}
A vector of realizations \(\mathbf{W}^{(i)}_\star\) obtained from a sequence of physical 
actions \(\mathcal{A}^{(i)}\) is said to be an i.i.d.\ sample from an \(\mathcal{L}_3\)-distribution 
\(Q = P(\mathbf{W}_\star)\) if \[P^\mathbb{C}(\mathbf{W}^{(i)}_\star = \mathbf{w} \mid \mathcal{A}^{(i)}) = P^{\mathcal{M}}(\mathbf{W}_\star = \mathbf{w})\quad \forall \mathbf{w},\]
for any structural causal model \(\mathcal{M} \in \mathcal{M}(\mathcal{G})\), 
where \(P^\mathbb{C}\) denotes the probability measure over the beliefs of the acting agent \(\mathbb{C}\).
\end{quote}

Using this notion of i.i.d.\ sampling, we now state the formal definition of realizability.

% Following~\citet{raghavan2025counterfactual}, the formal definition of realizability 
% can be given as follows:

\begin{quote}
\textbf{Definition (Realizability, Def. 3.4):}  
An $\mathcal{L}_3$-distribution \( Q = P(\mathbf{W}_\star) \) is \emph{realizable} given \( \mathcal{G} \) and \( \mathbb{A} \) if and only if there exists a sequence of actions \( \mathcal{A} \in \mathbb{A} \) such that the resulting vector of realizations \(\mathbf{W}_\star \) is an i.i.d. sample from \( P^\mathcal{M}(\mathbf{W}_\star) \) for any \(\mathcal{M} \in \mathcal{M}(\mathcal{G}) \).
\end{quote}

This goes beyond identifiability: while identifiability asks whether a quantity can be computed from observed data, realizability asks whether a quantity can be physically instantiated through feasible actions. A key constraint is the \textit{Fundamental Constraint of Experimentation (FCE)}:

\begin{quote}
\textbf{Assumption (FCE: Fundamental Constraint of Experimentation):}  
No unit in the target population may be subjected to the same causal mechanism \( f_V \in \mathcal{F} \) more than once.
\end{quote}

This constraint reflects the physical impossibility of re-subjecting the same unit to conflicting treatments (e.g., both \( X = x \) and \( X = x' \)). With this constraint in place, realizability can be fully characterized by the following criterion.

\paragraph{Realizability Criterion.}

\begin{quote}
\textbf{Corollary (Cor. 3.7):}  
An $\mathcal{L}_3$-distribution \( P(\mathbf{W}_\star) \) is realizable if and only if the ancestor set \( An(\mathbf{W}_\star) \) does not contain multiple versions of the same endogenous variable (e.g., \( W_t \) and \( W_s \)) under different regimes.
\end{quote}

This ensures that the necessary parent assignments for each potential response are not mutually exclusive---a prerequisite for jointly evaluating multiple counterfactuals on a single unit. Building on this theoretical foundation, we next present how realizability manifests in the design of our length bias mitigation strategy.

\paragraph{Application to Length Bias Mitigation.}

In our setting, each response \( T \) is modeled as a deterministic function of two latent variables: semantic content \( C \) and stylistic length preference \( L \), i.e., \( T = f(C, L) \). However, existing reward models tend to conflate the influence of \( C \) and \( L \), making it difficult to isolate the effect of one factor without confounding from the other. Importantly, the observed content variable \(C\) is not a purely semantic quantity: because it is learned jointly with stylistic factors, it often reflects an entangled representation that deviates from the true underlying meaning, which we denote by \(\tilde{C}\).

A naive strategy might attempt to intervene on \( L \) by modifying the input prompt—e.g., changing “What is an apple?” to “What is an apple? Answer in 5 words.” Let \( X \) denote the original prompt and \( X' \) the length-conditioned variant. Then the induced length can be seen as \( L_{x'} \), an intervention on \( L \) via prompt manipulation. However, this modification may also affect the true semantic content \( \tilde{C} \), both directly and indirectly. As a result, the observed content under prompt variation may not preserve the original semantics. Therefore, our objective is to reconstruct a disentangled representation that captures the core meaning of \( \tilde{C} \) while allowing targeted intervention on \( L \)—that is, a disentangled representation of content that is invariant to stylistic or length-based confounding.

To address this, our approach involves the controlled construction of counterfactual samples \( T_{\tilde{c}, \ell'} \) by explicitly intervening on both \( C \) and \( L \) via disjoint mechanisms:
\begin{itemize}
    \item \( L \) is set using prompt-based length conditioning;
    \item \( C \) is fixed by transplanting semantic content from a reference response;
    \item These two sources of intervention are applied independently, avoiding causal conflict.
\end{itemize}

For example, we might prompt the model with: \textit{“What is an apple? Answer in 5 words, while preserving the core meaning of <the original response>.”} A semantic consistency filter (e.g., a binary classifier) is then used to verify whether the generated response maintains the intended content.

This leads to the following counterfactual query:
\[
P\left( \text{R}(T_{\tilde{c}, \ell'}) \mid C = c, L = \ell \right),
\]
which asks: \textit{“What reward would the model assign if content were changed to \( \tilde{c} \) and length to \( \ell' \), given that the original sample had \( (C = c, L = \ell) \)?”}

Although the query appears to involve a change in content, we emphasize that \( \tilde{c} \) is constructed to preserve the semantic meaning of \( c \), independent of stylistic factors. Thus, \( \tilde{c} \) should not be interpreted as a semantically changed(different) version of \( c \), but rather as a disentangled reconstruction of its original, unconfounded meaning. The purpose of this query is not to modify content, but to isolate it from stylistic artifacts—enabling evaluation of the model’s reward under counterfactual variation in style while keeping semantics fixed.

This setting satisfies the realizability criterion of~\citet{raghavan2025counterfactual} (Corollary 3.7), which requires that no two versions of the same variable appear with conflicting parent assignments in the ancestor set \( \text{An}(\mathbf{W}_\star) \). Since \( C \) and \( L \) are manipulated through non-overlapping interventions and never overwritten simultaneously, our setup permits realizable counterfactual generation under the \textit{Fundamental Constraint of Experimentation (FCE)}.

\paragraph{Conclusion.}
Although $\mathcal{L}_3$ queries are generally not identifiable from observational or interventional data, they can be \emph{realized} under controlled environments that prevent conflicting causal assignments. By explicitly disentangling and intervening on content and length, we construct realizable counterfactuals that support robust and debiased training of reward models.

\section{Learning Length-Invariant Rewards via Counterfactuals}
\label{appendix:counterfactual-formalism}

We consider a generative setting where each response \( T \) is deterministically generated from a prompt \( X \), via latent factors: semantic content \( C \) and length style \( L \), i.e., \( T = f(C, L) \). The reward model \( R(X, T) \) maps prompts and completions to scalar rewards. However, since \( C \) and \( L \) are typically confounded in natural data, reward signals often entangle content quality with stylistic preferences such as verbosity.

Our objective is to train a reward model that is \textit{invariant} to length and \textit{sensitive} to semantic content. That is, rewards should reflect differences in content, not superficial stylistic variations. To achieve this, we employ a counterfactual augmentation strategy that simulates interventions on the latent factors \( C \) and \( L \). Specifically, we construct synthetic responses \( T_{\tilde{c}, \ell'} \) by:
\begin{itemize}
    \item Fixing the content \( C \leftarrow \tilde{c} \) via response transplantation from another instance;
    \item Controlling the length style \( L \leftarrow \ell' \) via prompt-level manipulation (e.g., stylistic constraints).
\end{itemize}
This yields counterfactual samples that decouple semantic meaning from stylistic artifacts, enabling more robust supervision.

\paragraph{Constructing counterfactual training pairs.}
We construct training examples where the length is held fixed while content varies, simulating content-based preferences under controlled stylistic settings.

Given an original dataset of the form
\[
\mathcal{D}_{\text{orig}} = \{(X, T_{c_1, \ell_1}, T_{c_2, \ell_2})\},
\]
we generate a counterfactual variant \( T_{\tilde{c}_1, \ell_2} \) that preserves the content of \( T_{c_1, \ell_1} \) but adopts the length of \( T_{c_2, \ell_2} \). This gives us a counterfactually aligned training pair:
\[
\mathcal{D}_{\text{cf}} = \{(X, T_{\tilde{c}_1, \ell_2}, T_{c_2, \ell_2})\}.
\]

By repeatedly training on such counterfactual pairs—where content is varied under fixed length—the reward model learns to prioritize semantic distinctions over stylistic ones. Over time, this reduces spurious correlations with verbosity and mitigates length bias in reward assignment. Therefore, we can train a reward model $R$ that satisfies:
\[
\underbrace{\sup_{T} \left[ \sigma(R(X, T_{c,\ell}) - R(X, T_{\tilde{c},\ell'}) ) \right]}_{\text{Effect of Length}}
\leq 
\underbrace{\inf_{T} \left[ \sigma(R(X, T_{c,\ell}) - R(X, T_{c',\ell}) ) \right]}_{\text{Effect of Content}} \quad \forall c \neq c', \; \ell \neq \ell'
\]
where \( \sigma(\cdot) \) is the sigmoid function used in pairwise preference modeling.

\section{Augmentation Techniques for Controlling Content and Length}
\label{appendix:augmentation-techniques}
This section describes our augmentation strategies to independently manipulate response length and semantic content. Each method is designed to manipulate either response verbosity or semantic information while controlling the other factor.

\subparagraph{Fixing content while changing length.}
To simulate \(T_{\tilde{c}, \ell'}\), we generate responses that preserve the core semantic content while varying the length bin. Specifically, we modify verbosity, style, or structure without introducing new substantive information. The following techniques enable controlled expansion or compression while maintaining meaning:

\begin{itemize}
    \item \textbf{Filler Sentences}: Inserting or removing sentences that serve to maintain flow or tone, without adding new information. 
    \textit{e.g.,} ``Here are the following reasons...'', ``It is important to note that...''

    \item \textbf{Pleonasm (Redundant Expression)}: Adding or deleting phrases that are semantically redundant within a sentence.  
    \textit{e.g.,} ``We are \textit{currently} in the \textit{present} moment.''

    \item \textbf{Redundant Sentences}: Repeating or removing information already stated to artificially modify length.  
    \textit{e.g.,} ``She was happy to see her dog. Meeting her dog was a pleasant moment.''

    \item \textbf{Paraphrasing / Summarization}: Restructuring sentences to compress or expand content while preserving meaning.  
    This includes active/passive transformations and sentence splitting or merging.  
    \textit{e.g.,} Merging two short sentences into one long sentence.

    \item \textbf{Format Changes}: Altering the presentation style or tone, such as converting prose into bullet points or shifting from academic to conversational language.  
    \textit{e.g.,} academic to conversational tone 
    % Academic: ``The results of the study indicate that there is a significant correlation between the two variables.''  
    % Conversational: ``The study shows that these two variables are closely related.''

    \item \textbf{Combination}: Using LLM-guided intuition to combine multiple augmentation strategies.  
    \textit{e.g.,} Filler + Paraphrasing, Redundancy + Format Change.
\end{itemize}

\subparagraph{Fixing length while changing content.}
To construct \(T_{c', \ell'}\), we generate semantically distinct responses that match the original length while modifying core content. The following techniques are used to vary meaning while keeping verbosity constant:

\begin{itemize}
    \item \textbf{Removing Necessary Details}: Omitting specific or descriptive elements while preserving the overall message.\\
    \textit{Before:} ``The new smartphone model, which was released last month after a series of delays and extensive testing, features an improved camera and a faster processor.''\\ 
    \textit{After:} ``The latest smartphone, launched last month following thorough evaluation, includes an upgraded camera and a more powerful processor''

    \item \textbf{Elaboration}: Adding examples or rephrasing to emphasize key points, while keeping the length constant. \\
    \textit{Before:} ``Studying improves cognitive skills.''\\  
    \textit{After:} ``Reading books enhances memory and focus.''

    \item \textbf{Information Substitution}: Replacing facts or details with alternative but same-type information. \\
     \textit{Before:} ``In 2020, global temperatures reached record highs due to climate change.''\\
    \textit{After:} ``In 2020, air pollution levels surged in major cities due to industrial activities.''

    \item \textbf{Converting Expression}: Rewriting figurative expressions into factual statements or vice versa.\\
     \textit{Before:} (Figurative): ``The negotiation was a game of chess, with each side making calculated moves.''\\  
    \textit{After:} (Factual): ``The negotiation involved careful strategy and analysis, with each decision impacting the outcome.''

    \item \textbf{Combination}: Generating diverse variations by blending multiple strategies. \textit{e.g.,} Elaboration + Information Substitution, or Removing Details + Converting Expression.
\end{itemize}

\section{Length Binning Criteria}
\label{appendix:length-bins}

\begin{table}[t]
\centering
\begin{tabular}{lc}
\toprule
\textbf{Category} & \textbf{Token Range} \\
\midrule
Very Short & (1, 41) \\
Short      & (41, 98) \\
Medium     & (98, 204) \\
Long       & (204, 377) \\
Very Long  & (377, 5220) \\
\bottomrule
\end{tabular}
\caption{Length binning criteria for responses.}
\label{tab:binning}
\end{table}

As illustrated in ~\Cref{tab:binning}, we discretize response lengths into five bins to define "fixed length" and "different length" conditions, which are essential for counterfactual data augmentation and bias diagnosis. This binning allows us to simulate length-based interventions while maintaining consistency in how we interpret \textit{length equivalence} during training and evaluation. The bins were selected based on quantiles from the empirical distribution of response lengths in the training corpus. Responses falling within the same bin are treated as approximately equal in length, while response pairs that fall into different bins are considered to have length divergence. This coarse binning scheme provides a balance between granularity and robustness: it prevents small token-level differences from being interpreted as meaningful length variations, while enabling us to capture broad shifts in verbosity.

\begin{figure}[H]
\centering
\includegraphics[width=0.6\linewidth]{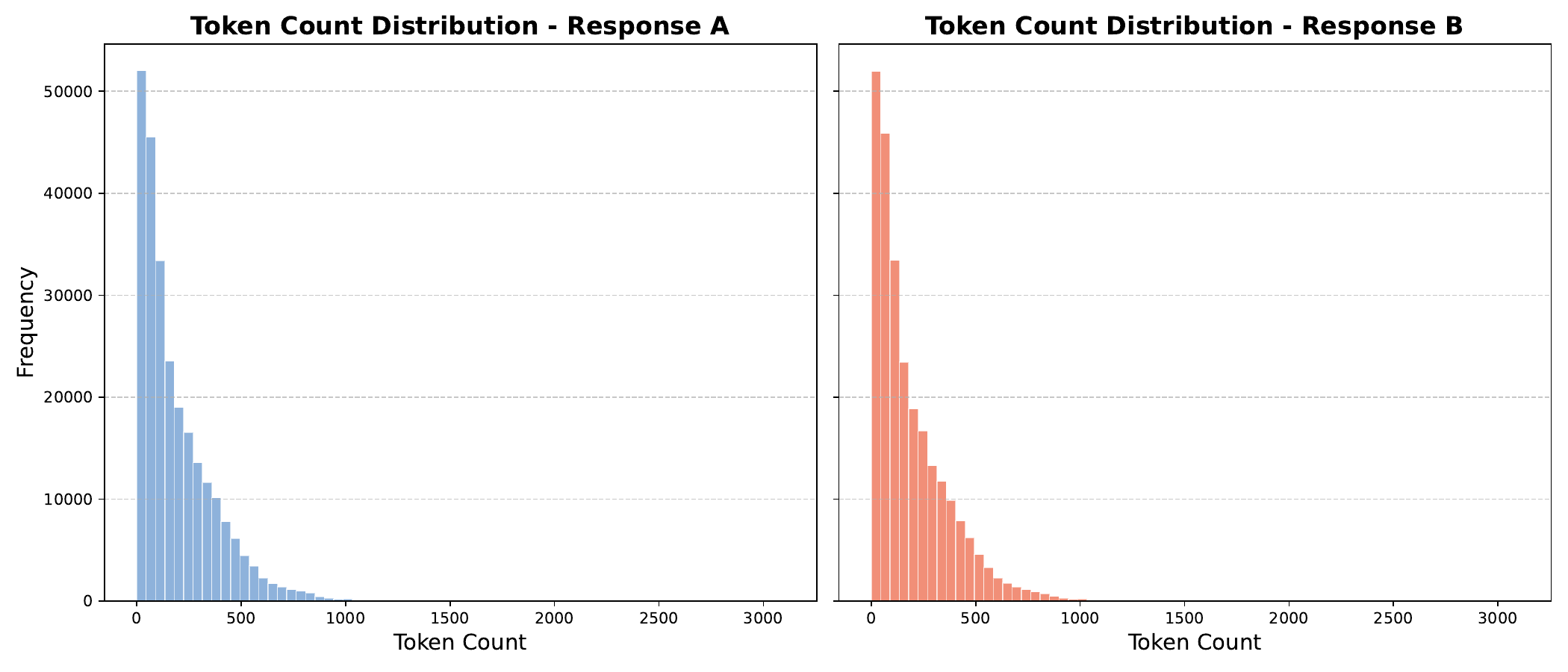}
\caption{Token count distributions for Response~A (left) and Response~B (right) in the training corpus.}
\label{fig:token-distribution}
\end{figure}

One might question whether the last bin (Very Long: 377--5220 tokens) is too wide to be considered a single category. However, as shown in~\Cref{fig:token-distribution}, such extreme-length responses are rare. Specifically, only about 5.6\% of responses exceed 512 tokens (12,982 in Response A and 12,936 in Response B, out of 230,427 total), indicating that the long tail of the distribution consists primarily of statistical outliers. Therefore, the width of the final bin does not undermine the validity of our length control strategy in practice.

\section{Augmentation Strategies}
\label{appendix:augmentation-examples}

\Cref{tab:prompt-content} and \Cref{tab:prompt-length} show the specific prompts we used for augmentation in two directions: (1) fixing semantic content while varying length, and (2) fixing length while varying content. 
Applying these prompts to the same original completion yields the augmented responses in \Cref{tab:fix-content} and \Cref{tab:fix-length}.
All augmented responses are derived from the following original completion:

\begin{table}[H]
\centering
\begin{tabular}{m{0.482\linewidth} m{0.482\linewidth}}
\includegraphics[width=\linewidth]{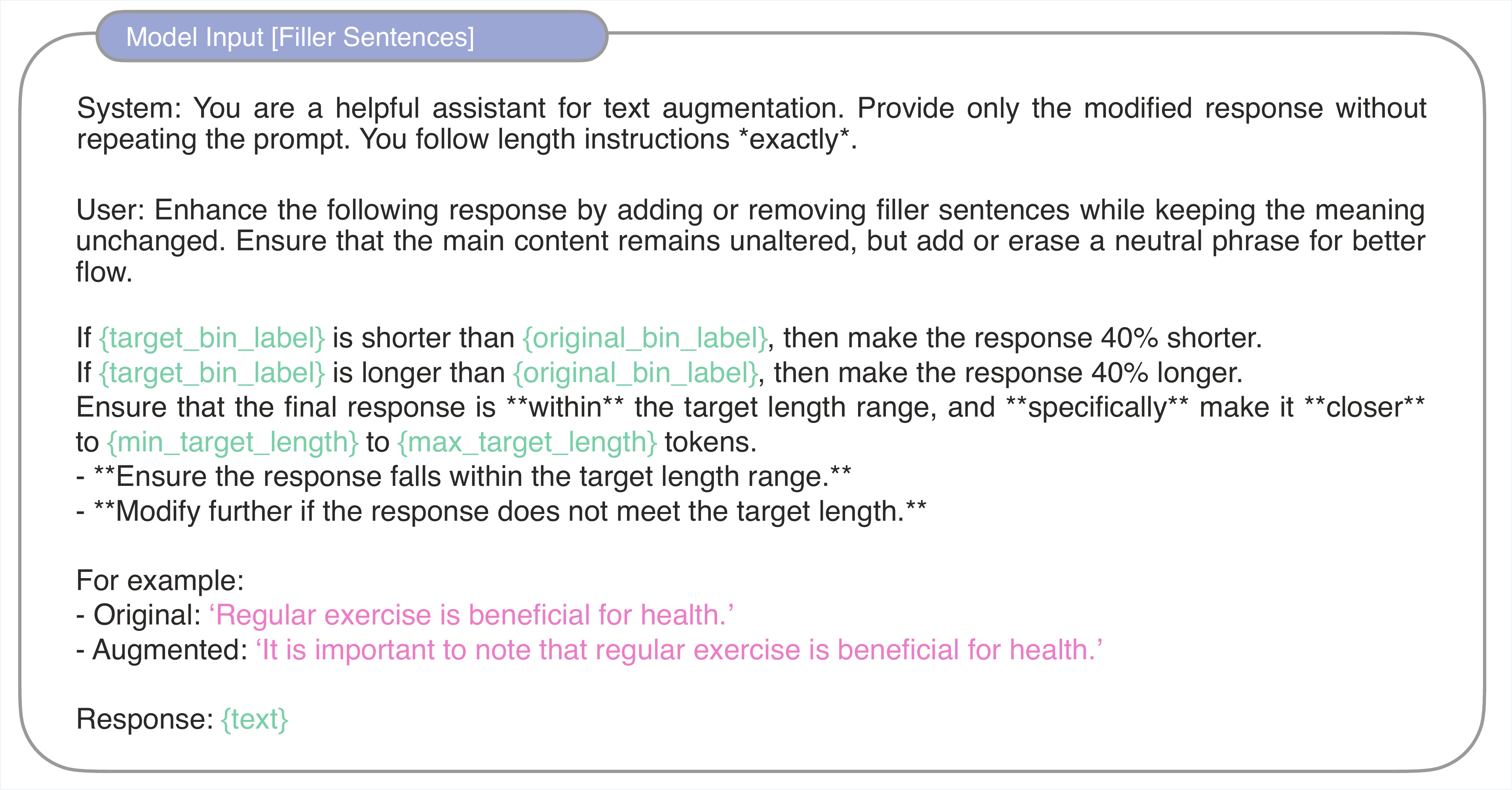} &
\includegraphics[width=\linewidth]{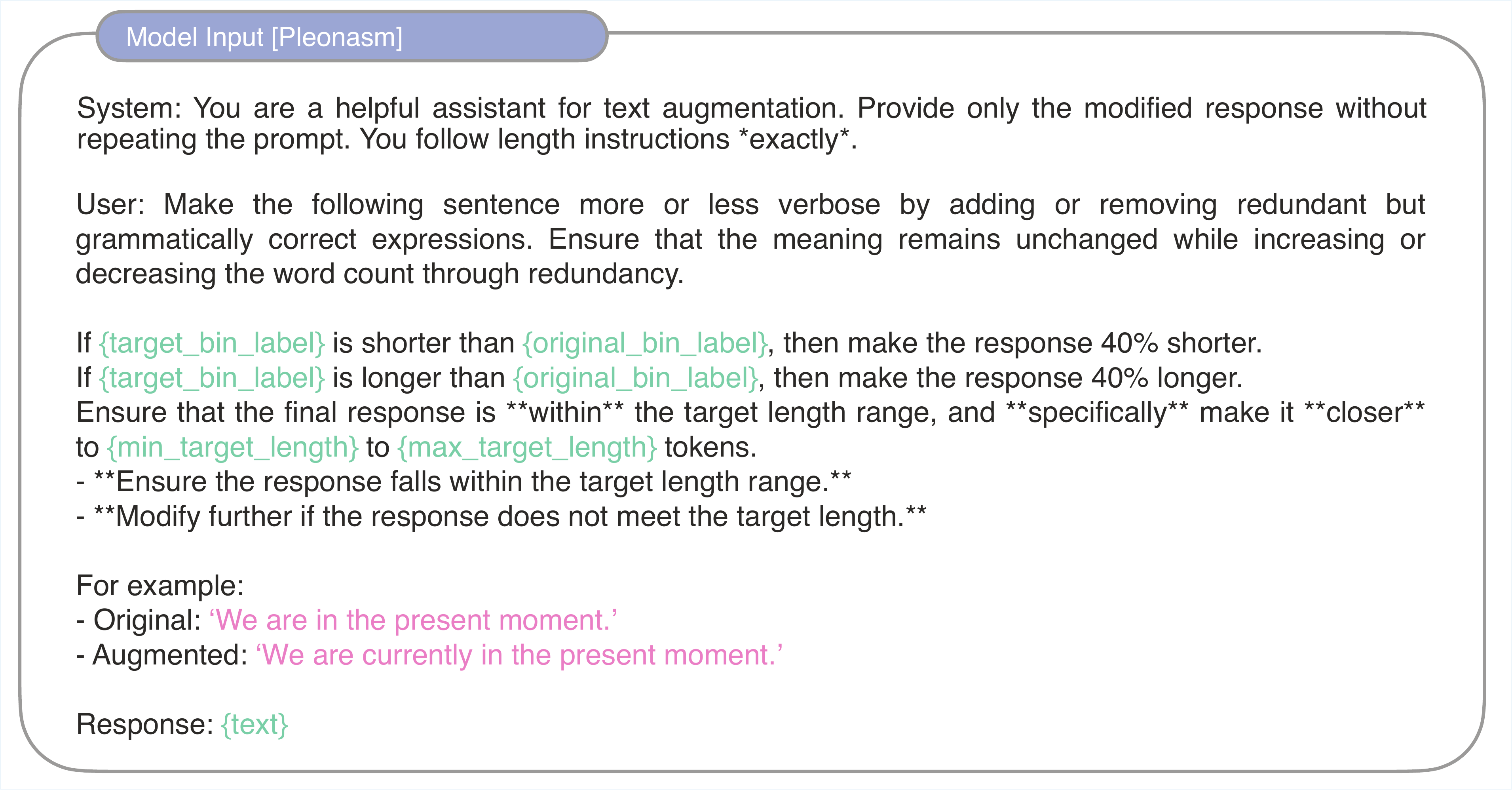} \\[1.5em]
\includegraphics[width=\linewidth]{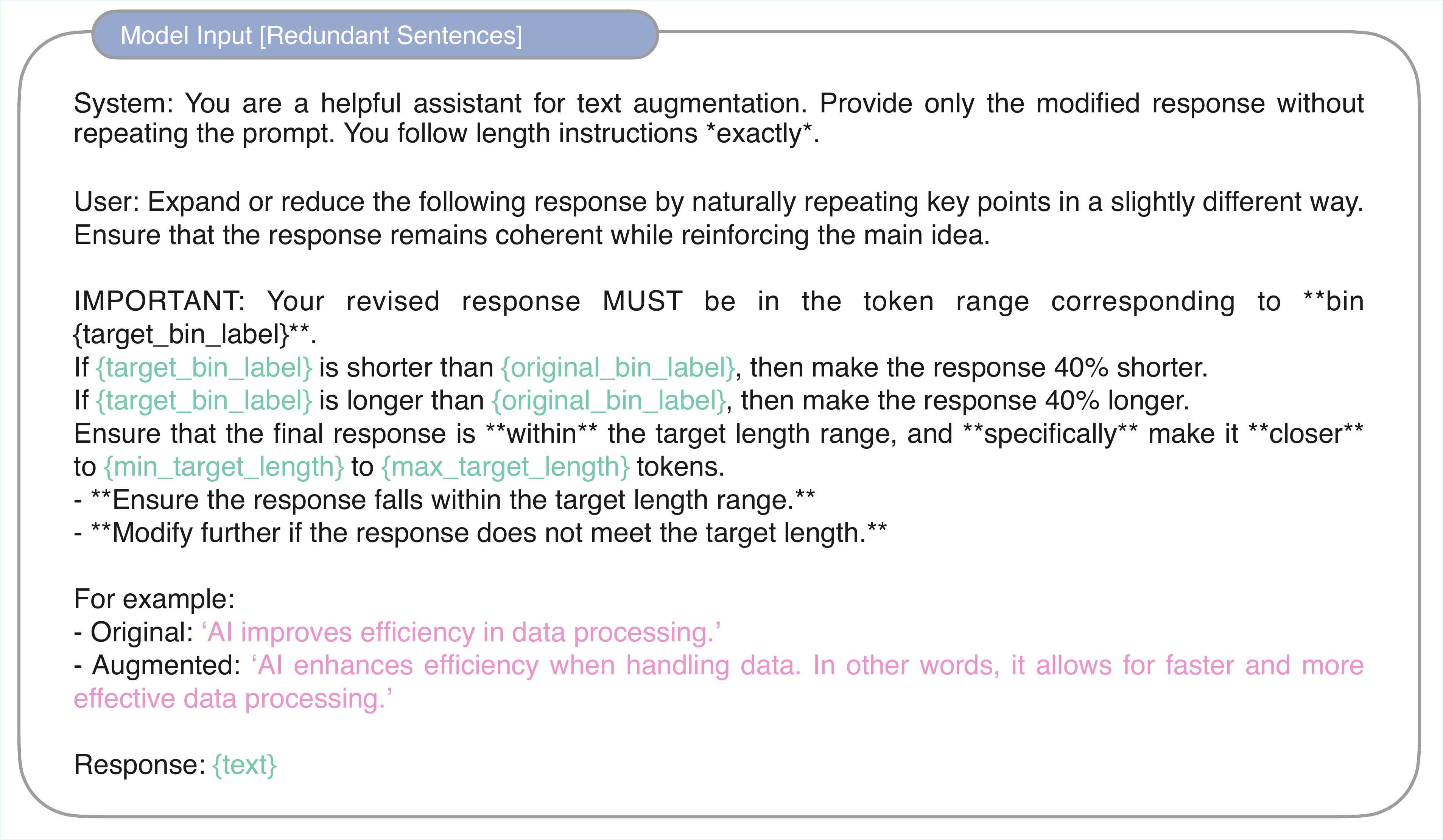} &
\includegraphics[width=\linewidth]{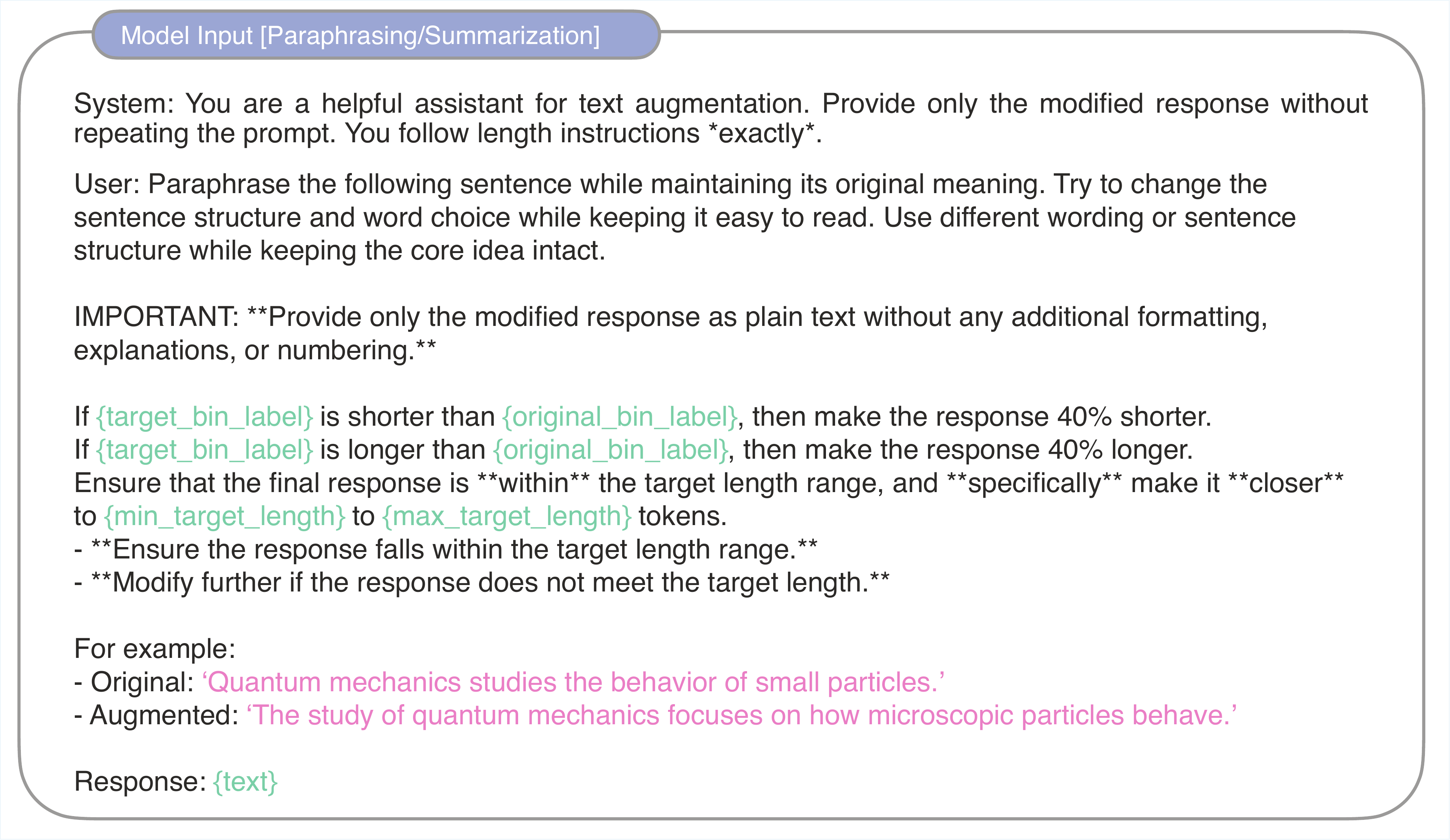} \\[1.5em]
\includegraphics[width=\linewidth]{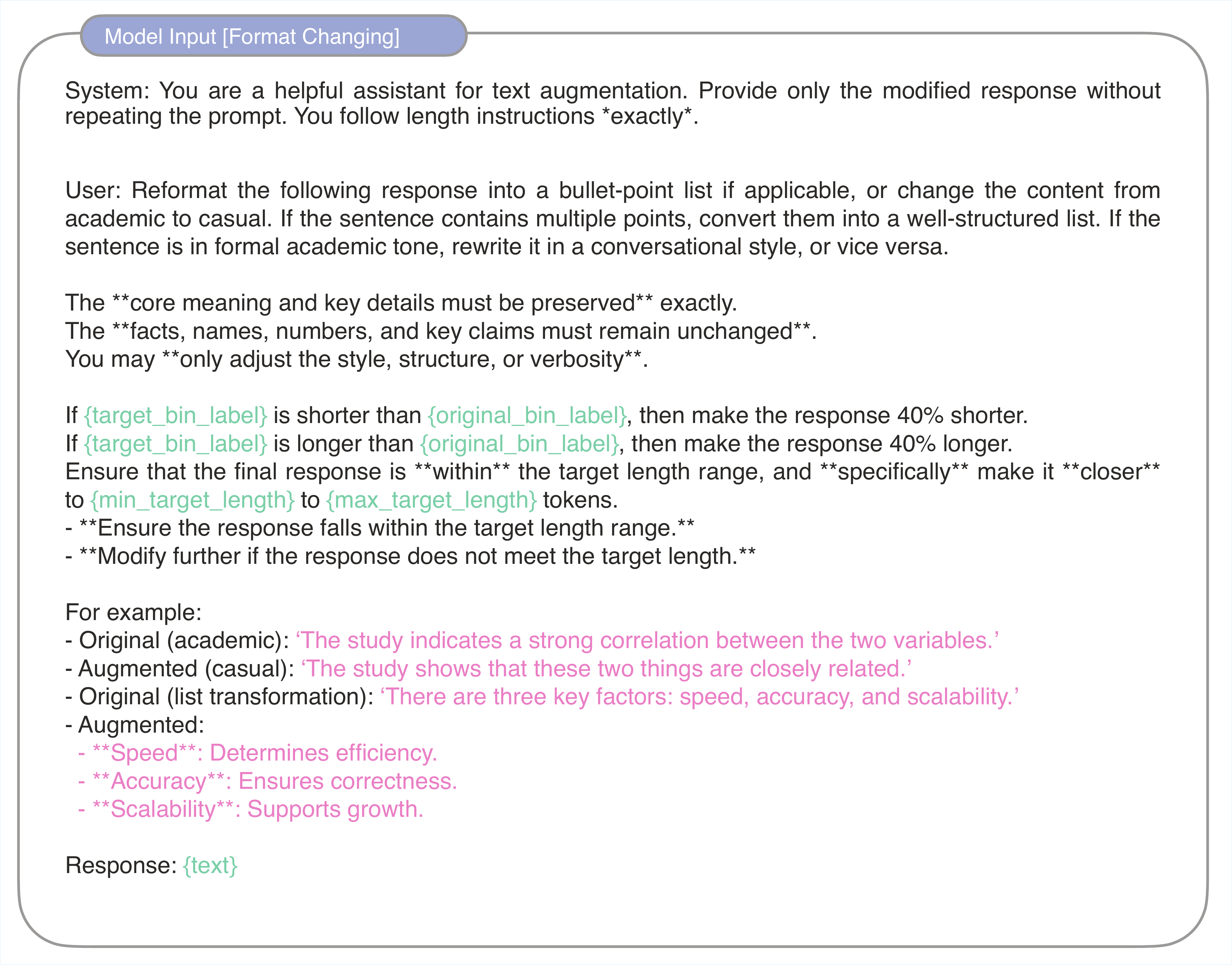} &
\includegraphics[width=\linewidth]{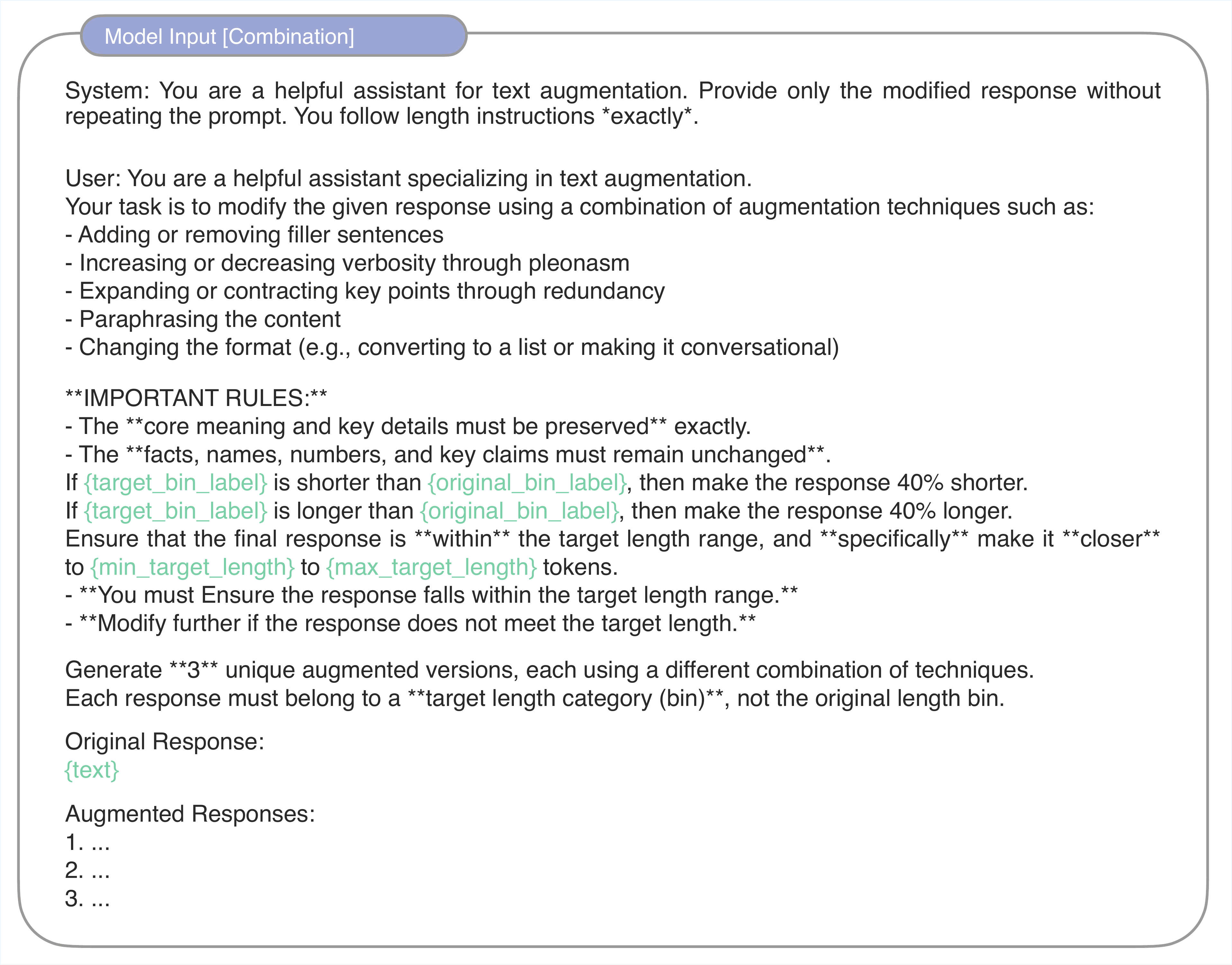} \\[1.5em]
\end{tabular}
\caption{Prompt for fixing semantic content while varying length.
Each method applies a different augmentation strategy while preserving the original meaning.}
\label{tab:prompt-content}
\end{table}

\begin{table}[H]
\centering
\begin{tabular}{m{0.482\linewidth} m{0.482\linewidth}}
\includegraphics[width=\linewidth]{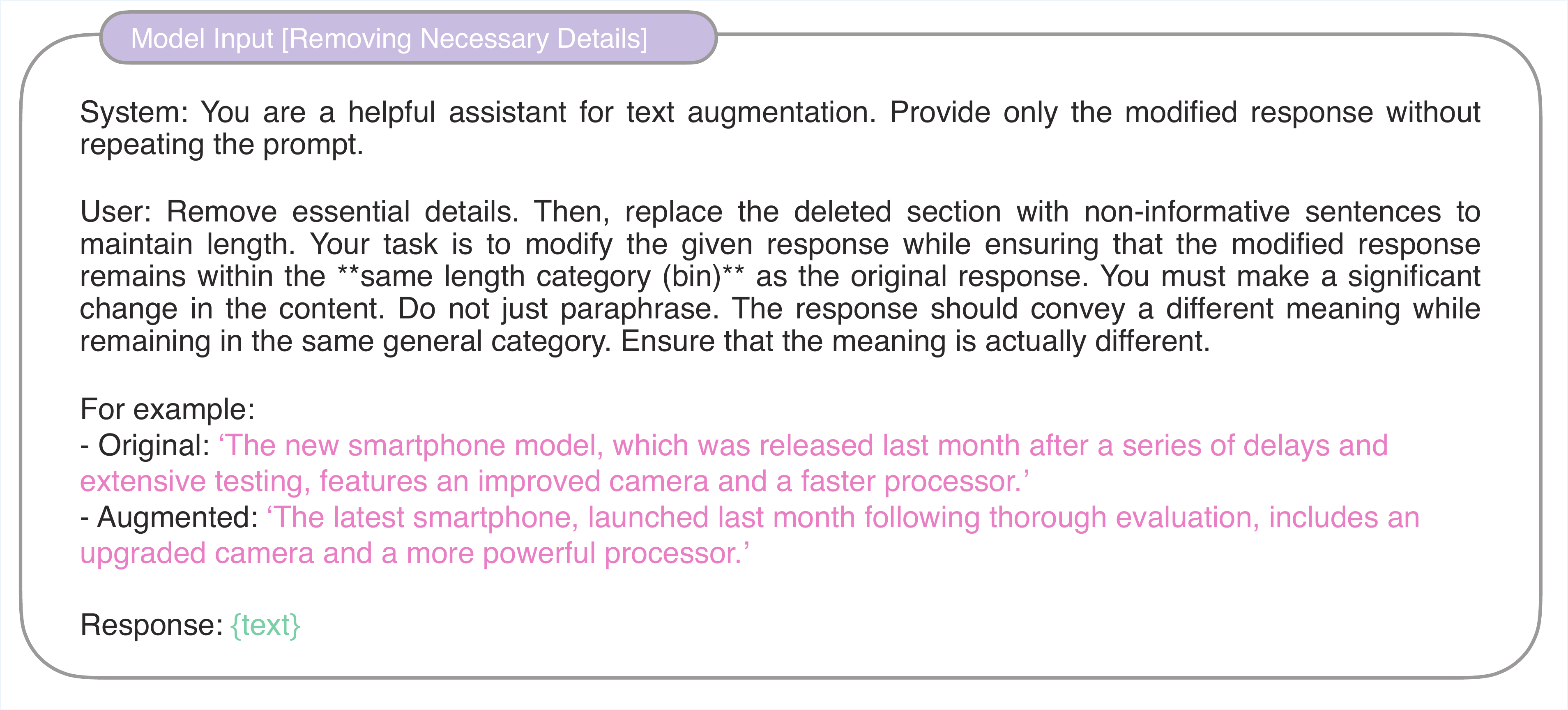} &
\includegraphics[width=\linewidth]{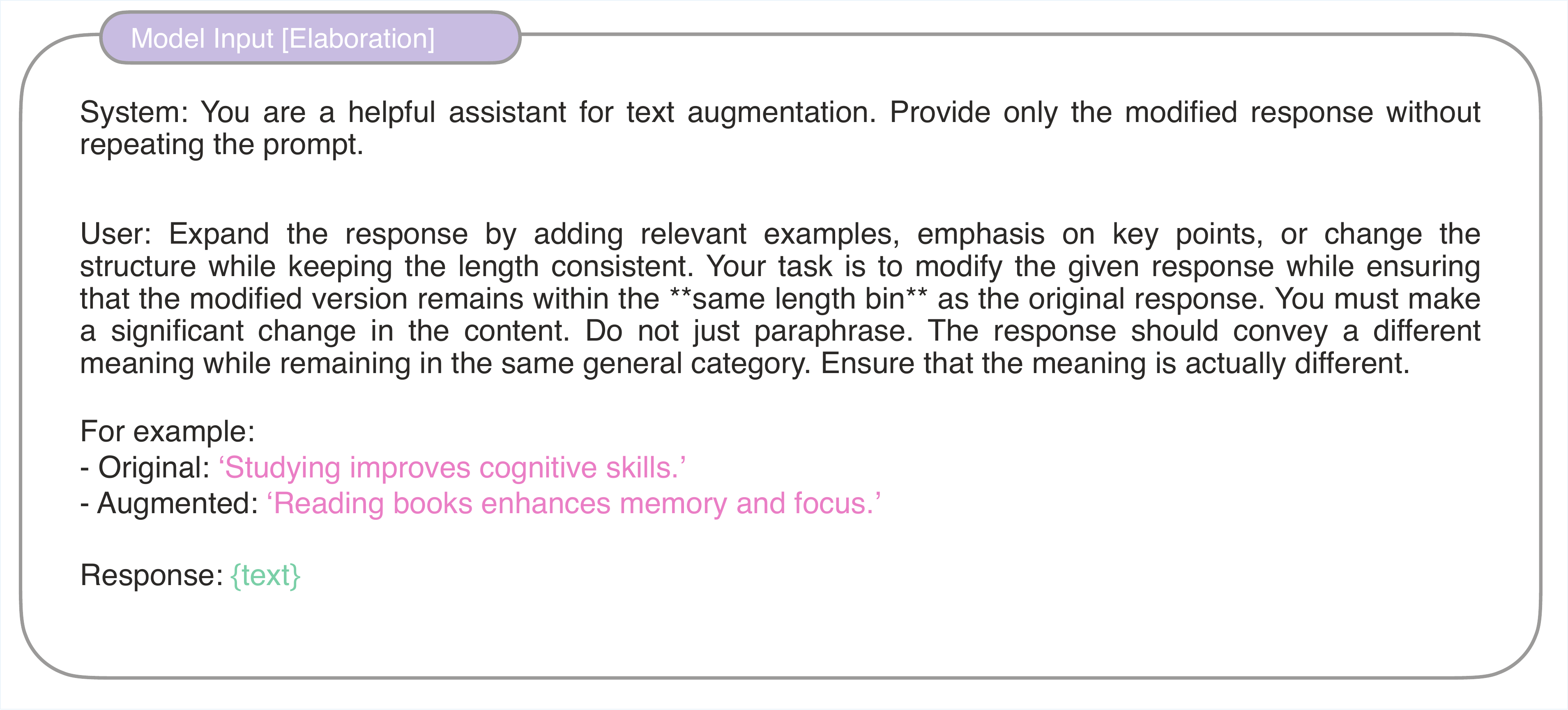} \\[1.5em]
\includegraphics[width=\linewidth]{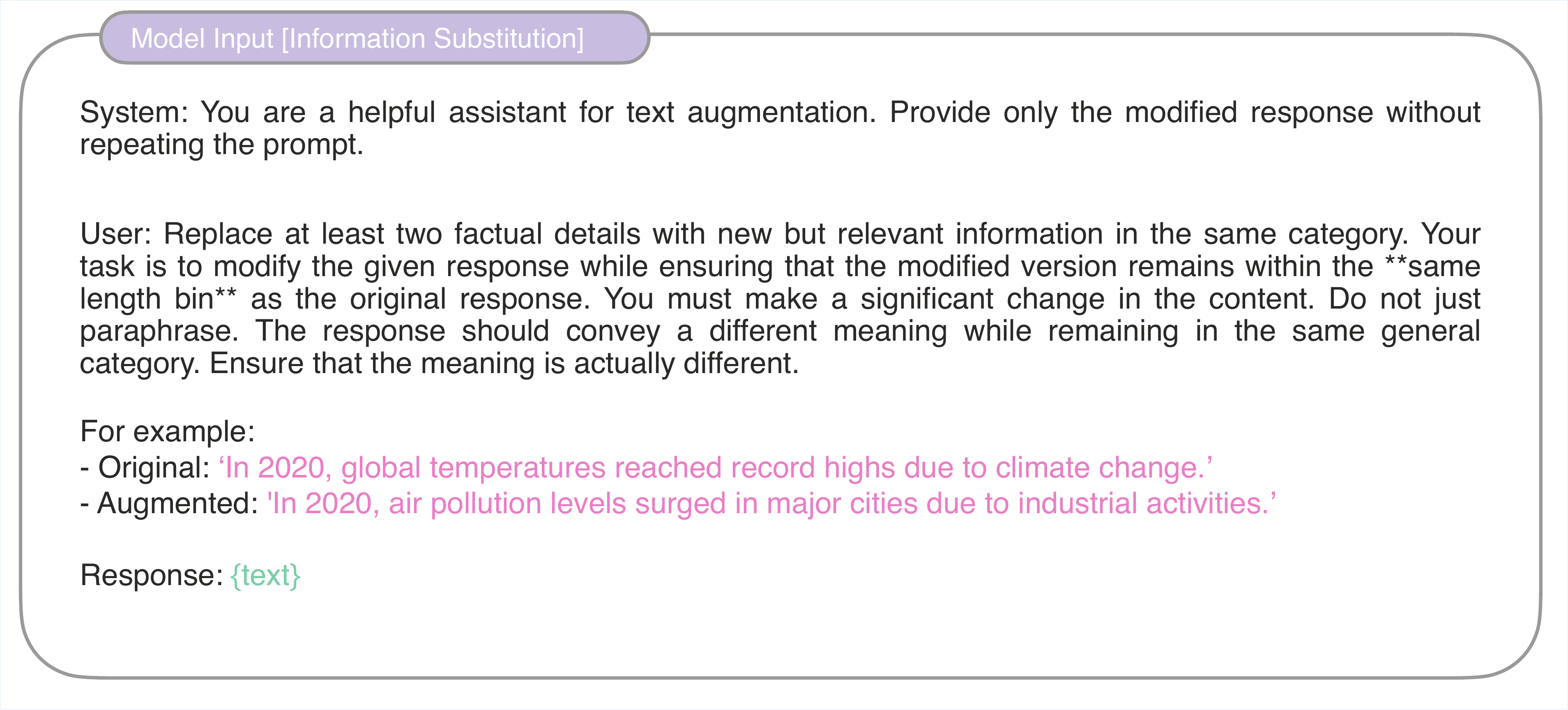} &
\includegraphics[width=\linewidth]{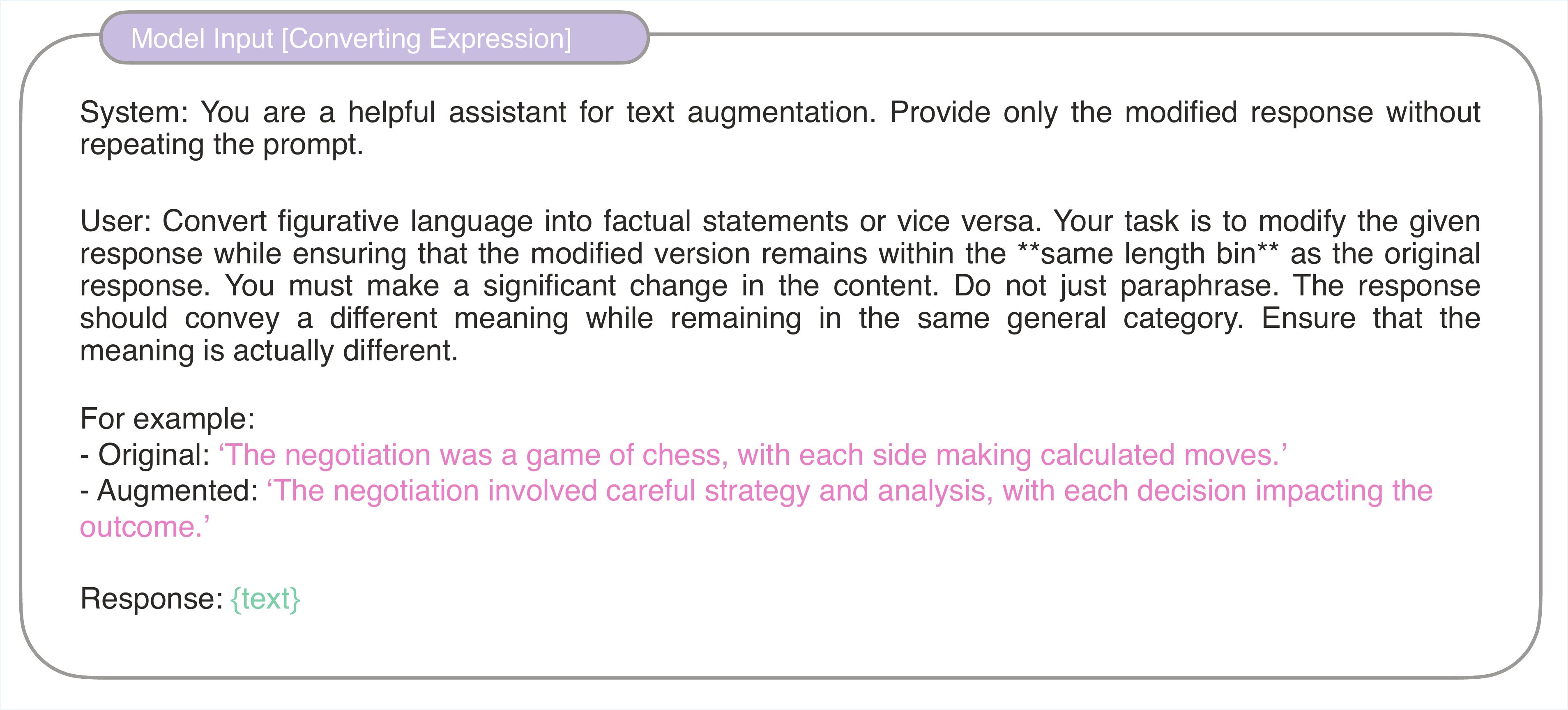} \\[1.5em]
\multicolumn{2}{c}{
\includegraphics[width=0.55\linewidth]{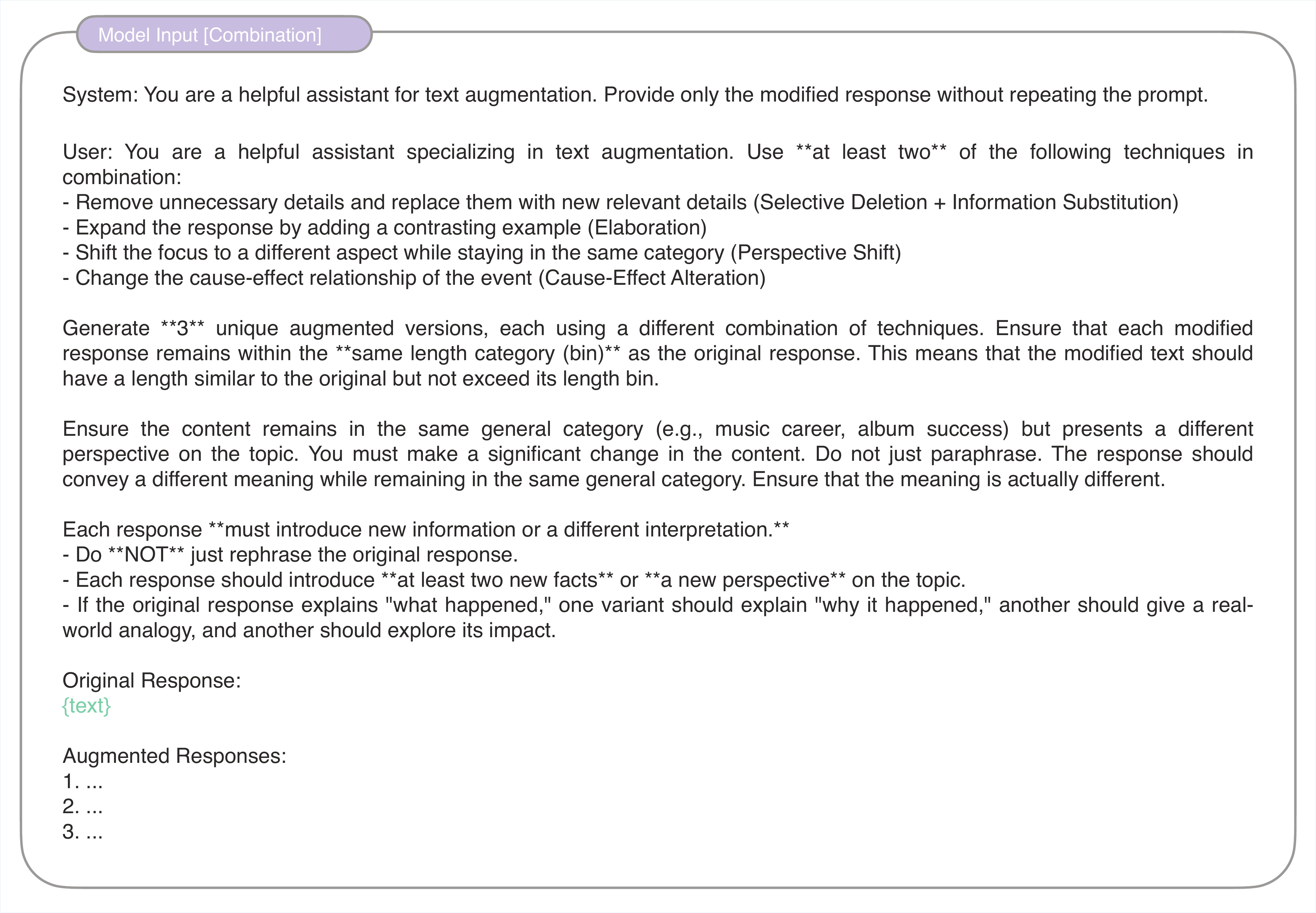}
} \\[1.5em]
\end{tabular}
\caption{Prompt for varying semantic content while keeping length fixed.
Each method applies a different augmentation strategy while keeping the original length bin.}
\label{tab:prompt-length}
\end{table}

% \Cref{tab:fix-content} and \Cref{tab:fix-length} provide concrete examples used for augmenting responses in two directions: (1) fixing semantic content while varying length, and (2) fixing length while varying content. All augmented responses are derived from the following original completion:
The following tables (\Cref{tab:fix-content} and \Cref{tab:fix-length}) present the augmented responses generated by applying the prompts from the previous figures. 
Each method illustrates a distinct augmentation strategy within the corresponding constraint (fixed content or fixed length).

\noindent
\textbf{Original Response:} \\
\textit{You're right, it was his fourth studio album. He also had major success with the album The Stranger, and his newer album River of Dreams. He's been active since the 1970s and is still active today, and his albums were also some of the best-selling solo albums ever released.}

\begin{table}[H]
\scriptsize
\centering
\begin{adjustbox}{width=\textwidth}
\begin{tabular}{|p{3.5cm}|p{10.5cm}|}
\hline
\textbf{Method} & \textbf{Augmented Response} \\
\hline

Filler Sentences & You're absolutely right, \textit{it is indeed true} that it was his \dots. \textit{In addition to that}, he also had \dots. \textit{It's fascinating to think} that he has been active \dots. \textit{Furthermore, it's worth mentioning} that his albums were also \dots. \\
\hline

Pleonasm (Redundant Expression) & You are absolutely correct, it was indeed his fourth studio album. Additionally, he achieved \dots The Stranger, \textit{as well as} with his more recent album \dots industry since the \textit{decade of the 1970s} and \dots have ever been \textit{released to the public}. \\
\hline

Redundant Sentences & You're absolutely correct; it was indeed \dots. It's worth noting that his albums rank among the best-selling solo albums in music history, \textit{showcasing his enduring popularity and influence}. \\
\hline

Paraphrasing / Summarization & You are correct; it was his fourth studio album. He achieved significant success with The Stranger and his more recent album, River of Dreams. Active since the 1970s, he continues to work today, and his albums are among the best-selling solo records. \\
\hline

Format Changing & 
\begin{tabular}[t]{@{}l@{}}
-- You're spot on! It was his fourth studio album. \\
-- He hit it big with: \\
\quad \textbf{The Stranger}: One of his major successes. \\
\quad \textbf{River of Dreams}: His newer album that's also done well. \\
-- He kicked off his career in the 1970s and is still going strong. \\
-- Plus, his albums are among the best-selling solo albums ever!
\end{tabular} \\
\hline

Combination & Absolutely, you are spot on! Indeed, that particular album \dots alongside his more contemporary release, River of Dreams. \textit{It's quite remarkable} that he has maintained \dots prolific even today. \textit{Not to mention}, many of his albums stand out \dots. \\
\hline
\end{tabular}
\end{adjustbox}
\caption{Examples of Fixing Content while Varying Length.}
\label{tab:fix-content}
\end{table}

\begin{table}[H]
\scriptsize
\centering
\begin{adjustbox}{width=\textwidth}
\begin{tabular}{|p{3.5cm}|p{10.5cm}|}
\hline
\textbf{Method} & \textbf{Augmented Response} \\
\hline
Removing necessary details & \dots he achieved notable acclaim \textit{with a different project}, and \textit{his recent endeavors} have garnered attention. Since the \textit{late 20th century}, he continues to engage \dots, and his works have been \textit{widely recognized} across different mediums. \\
\hline
Elaboration & Absolutely, it was indeed his fourth studio album. Additionally, \dots hit singles like \textit{"Just the Way You Are"} from The Stranger and \textit{"All About Soul"} from River of Dreams \dots stand out as some of the \textit{top-selling records} of all time, showcasing his \textit{enduring influence and popularity in the industry}. \\
\hline
Information Substitution & You're correct, it was his \textit{fifth} studio album. He also achieved \dots album \textit{Glass Houses}, and his latest release is titled \textit{My Lives}. He has been performing since the \textit{1980s} and \dots ranked among the \textit{highest-grossing live albums} in history. \\
\hline
Converting Expression & \dots His career began in the 1970s, and he has since \textit{retired from the music scene}, with his albums now considered \textit{classic hits in rock history}. \\
\hline
Combination & Indeed, while it was his fourth studio album, his earlier work \textit{laid the foundation for his distinctive sound}. In particular, "The Stranger" showcased his \textit{storytelling ability}, while "River of Dreams" reflected his \textit{exploration of personal themes}. \dots \textit{his focus has also shifted toward mentoring new artists}, significantly impacting the music scene. \\
\hline
\end{tabular}
\end{adjustbox}
\caption{Examples of Fixing Length while Varying Content.}
\label{tab:fix-length}
\end{table}

\section{Fine-Tuning Details}
\label{appendix:finetuning}
This appendix provides implementation details for all fine-tuning steps used in the paper, including classification filtering, reward modeling, and reinforcement learning. All fine-tuning experiments were conducted on a Supermicro 740GP-TNRT server with dual Xeon 6342 CPUs and four NVIDIA A6000 GPUs.

\subsection{Cross-Encoder Fine-Tuning}
\label{appendix:crossencoder-finetuning}

We fine-tune a cross-encoder classifier to distinguish between content-preserving and content-altering response pairs for semantic filtering. The model is based on the \texttt{All-Mpnet-Base-V2} architecture from SentenceTransformers, with a single regression head for binary classification. The output is a single scalar logit per pair, passed through a sigmoid to estimate semantic equivalence probability.

\paragraph{Data.} We use approximately 945K examples consisting of content-fixed and length-fixed augmentations. Content-fixed pairs are labeled as $1.0$ (semantically equivalent), while length-fixed pairs are labeled as $0.0$ (semantically divergent). A 95/5 train-validation split is applied.

\paragraph{Preprocessing.} 
We tokenize each response pair using \texttt{AutoTokenizer}, concatenating the prompt with the chosen and rejected responses. Inputs are truncated to a maximum length of 512 tokens using the \texttt{longest\_first} strategy and dynamically padded within each batch via \texttt{DataCollatorWithPadding}, all implemented through the Hugging Face Transformers library~\citep{wolf2020transformers}.

\paragraph{Training.} We fine-tune the model using a custom Hugging Face \texttt{Trainer} implementation with binary cross-entropy loss. Optimization is performed using AdamW with weight decay, and mixed precision (FP16) is enabled. Full hyperparameter details are provided in Table~\ref{tab:cross_encoder_training}.

% \paragraph{Training.} The model is fine-tuned using the Hugging Face \texttt{Trainer} API with the following configuration:
% \begin{itemize}
%   \item Learning rate: $2 \times 10^{-5}$
%   \item Batch size: 32
%   \item Epochs: 3
%   \item Optimizer: AdamW with weight decay $= 0.01$
%   \item Loss: Binary cross-entropy over sigmoid-scaled logits
%   \item Mixed precision (FP16) enabled
%   \item Best checkpoint selected by lowest validation loss
% \end{itemize}
% The main hyperparameters are shown in Table~\ref{tab:cross_encoder_training}.

\begin{table}[ht]
\centering
\begin{tabular}{p{0.22\linewidth} p{0.38\linewidth}}
    \hline
    \textbf{Hyperparameter} & \textbf{Value} \\
    \hline
        Learning rate           & $2 \times 10^{-5}$ \\
        Batch size              & 32 \\
        Epochs                  & 3 \\
        Optimizer               & AdamW (weight decay = 0.01) \\
        Loss                    & Binary cross-entropy over sigmoid-scaled logits \\
        Precision               & Mixed precision (FP16) \\
        Checkpoint selection    & Lowest validation loss \\
    \hline
\end{tabular}
\caption{Training configuration for cross-encoder fine-tuning.}
\label{tab:cross_encoder_training}
\end{table}

\paragraph{Filtering.} At inference time, the fine-tuned cross-encoder assigns a probability score to each response pair. For content-fixed pairs, we retain those with predicted probability \( p \geq 0.5 \), while for length-fixed pairs, we retain those with \( p < 0.5 \). Inference is performed in batches of 256 on a GPU, and the filtered outputs are stored in JSONL format.

\subsection{Reward Model Fine-tuning Details}
\label{appendix:rm-finetuning}
This section describes the fine-tuning setups for the reward models used in our experiments: \texttt{CDA\_OpenLM}, \texttt{CDA\_LoRA}, and \texttt{CDA\_HRO}.

\paragraph{Preprocessing.} 
We use a counterfactually augmented dataset of \texttt{(prompt, chosen, rejected)} triplets. The dataset is shuffled with a fixed seed and split into 95\% training and 5\% evaluation subsets. Each input is tokenized by concatenating the prompt with both responses and truncated to a maximum length of 512 tokens. Dynamic padding is applied within each batch, and inputs are formatted as PyTorch tensors for pairwise ranking.

% \begin{table}[H]
% \centering
% \begin{tabular}{p{0.25\linewidth} p{0.4\linewidth}}
% \hline
% \textbf{Setting} & \textbf{Value} \\
% \hline
%     Maximum sequence length & 512 tokens \\
%     Padding & Applied to the maximum length within each batch or fixed-length padding \\
%     Truncation & Enabled to truncate sequences longer than the maximum length \\
%     Padding token & EOS token if no pad token is available \\
% \hline
% \end{tabular}
% \caption{Tokenization settings for reward model fine-tuning}
% \label{tab:rm_tokenization}
% \end{table}

\paragraph{Reward Model Architectures.}
\begin{itemize}
    \item \textbf{CDA\_OpenLM:}  
    Initialized from the LLaMA-3B-based sequence classification model \texttt{openlm-research/open\_llama\_3b}\footnote{\url{https://huggingface.co/openlm-research/open_llama_3b}}, which outputs a scalar reward score.
    
    \item \textbf{CDA\_LoRA:}  
    Applies LoRA adapters~\citep{hu2021lora} to the query, key, and value projections of the base model \texttt{weqweasdas/hh\_rlhf\_rm\_open\_llama\_3b}\footnote{\url{https://huggingface.co/weqweasdas/hh_rlhf_rm_open_llama_3b}}, enabling parameter-efficient fine-tuning.

    \item \textbf{CDA\_HRO:}  
    Fully fine-tuned version of the same architecture as \texttt{CDA\_LoRA}, trained on our counterfactually augmented dataset.
\end{itemize}

\paragraph{Training.}
All models are fine-tuned using a pairwise ranking objective optimized with the margin ranking loss:
\[
\mathcal{L} = \max\left(0, m - s_{\text{chosen}} + s_{\text{rejected}}\right),
\]
where \( s_{\text{chosen}} \) and \( s_{\text{rejected}} \) are the scalar reward scores assigned to the chosen and rejected responses, and \( m = 0.5 \) is the margin hyperparameter. Training is performed using the Hugging Face \texttt{Trainer} API, with mixed-precision (bfloat16) enabled and automatic device placement. A custom data collator is used to pad the chosen and rejected sequences independently. Evaluation is conducted every 5,000 or 10,000 steps depending on the setup, and early stopping is applied based on validation loss. LoRA models are implemented using the PEFT library~\citep{peft2023github} with a rank \( r = 16 \), scaling factor \( \alpha = 32 \), and dropout rate of 0.01.

% Training hyperparameters are as follows:
% \begin{itemize}
%     \item Optimizer: AdamW
%     \item Learning rate: \(1 \times 10^{-5}\)
%     \item Batch size: 1 for full fine-tuning, 2 for LoRA fine-tuning
%     \item Number of epochs: 3
%     \item Mixed precision training with bfloat16 (BF16) enabled
%     \item Gradient accumulation steps: 2–4 depending on setup
%     \item Learning rate scheduler: cosine decay with linear warmup (5\% of total steps)
% \end{itemize}

% The main hyperparameters are summarized in Table~\ref{tab:rm_training_config}.

\begin{table}[ht]
\centering
\begin{tabular}{p{0.25\linewidth} p{0.38\linewidth}}
\hline
\textbf{Hyperparameter} & \textbf{Value} \\
\hline
Optimizer & AdamW \\
Learning rate & \(1 \times 10^{-5}\) \\
Batch size & 1 (full fine-tuning), 2 (LoRA fine-tuning) \\
Number of epochs & 3 \\
Precision & bfloat16 (BF16) \\
Gradient accumulation steps & 2–4 (depending on model size) \\
Learning rate scheduler & Cosine decay with linear warmup (5\% of steps) \\
Evaluation frequency & Every 5k–10k steps \\
Early stopping & Based on validation loss \\
\hline
\end{tabular}
\caption{Training configuration for reward model fine-tuning.}
\label{tab:rm_training_config}
\end{table}

\paragraph{Evaluation.}
Model performance during training is monitored via pairwise accuracy on the validation set, calculated as the fraction of examples where \(s_{\text{chosen}} > s_{\text{rejected}}\). Final checkpoint models are saved for downstream evaluation on RewardBench and length-controlled accuracy metrics. See~\Cref{appendix:alpacaeval} for further evaluation details.

\subsection{Supervised Fine-Tuning(SFT) Details}
\label{appendix:sft}
This section describes the supervised fine-tuning (SFT) setup used for initializing the policy model before reinforcement learning.

\paragraph{Preprocessing.}
We use the ShareGPT dataset~\citep{zheng2023judging}, which consists of prompt–response pairs derived from real-world dialogue. Each example is formatted for causal language modeling as:
\begin{quote}
\texttt{<prompt>\textbackslash n\textbackslash n<response><eos>}
\end{quote}
Tokenization is performed with the tokenizer from \texttt{openlm-research/open\_llama\_3b}, applying a maximum sequence length of 512 tokens. Sequences are truncated as needed and padded using the EOS token. The data is split into 95\% training and 5\% evaluation sets. Gradient checkpointing is enabled to reduce memory usage.

\paragraph{Training.}
We fine-tune the policy model with standard causal language modeling objectives using the Hugging Face \texttt{Trainer} API. Optimization is performed with the Adam8bit optimizer from \texttt{bitsandbytes}, and training runs in mixed precision (FP16). Automatic device placement via \texttt{device\_map="auto"} allows efficient utilization of available multi-GPU hardware. Table~\ref{tab:sharegpt_finetuning} summarizes the full training configuration.

% \begin{itemize}[itemsep=2pt, topsep=2pt]
%     \item Model: \texttt{open\_llama\_3b}
%     \item Tokenizer padding token: EOS
%     \item Max sequence length: 512 tokens
%     \item Optimizer: Adam8bit (bitsandbytes)
%     \item Learning rate: $5 \times 10^{-6}$
%     \item Weight decay: 0.01
%     \item Warmup ratio: 0.1
%     \item Batch size: 1 (per device)
%     \item Epochs: 2
%     \item Evaluation steps: every 10{,}000 steps
%     \item Save steps: every 10{,}000 steps
%     \item Precision: \texttt{fp16}
%     \item Gradient checkpointing: Enabled
%     \item Data collator: \texttt{DataCollatorForLanguageModeling (mlm=False)}
% \end{itemize}
\begin{table}[ht]
\centering
\begin{tabular}{p{0.25\linewidth} p{0.38\linewidth}}
\hline
\textbf{Hyperparameter} & \textbf{Value} \\
    \hline
    Model & \texttt{open\_llama\_3b} \\
    Tokenizer padding token & EOS \\
    Maximum sequence length & 512 tokens \\
    Optimizer & Adam8bit (\texttt{bitsandbytes}) \\
    Learning rate & $5 \times 10^{-6}$ \\
    Weight decay & 0.01 \\
    Warmup ratio & 0.1 \\
    Batch size (per device) & 1 \\
    Epochs & 2 \\
    Evaluation steps & Every 10{,}000 steps \\
    Save steps & Every 10{,}000 steps \\
    Precision & FP16 \\
    Gradient checkpointing & Enabled \\
    Data collator & \texttt{DataCollatorForLanguageModeling} (mlm=False) \\
    \hline
\end{tabular}
\caption{Fine-tuning configuration for the ShareGPT dataset on \texttt{open\_llama\_3b}.}
\label{tab:sharegpt_finetuning}
\end{table}

The resulting fine-tuned model is used to initialize all RLHF policy models trained via PPO.

\paragraph{Policy models used in RLHF evaluation.}
\label{appendix:policy-models}
Starting from the SFT model, we trained three RLHF policy models using PPO: \textbf{PPO\_HRO}, \textbf{PPO\_CDA\_OpenLM}, and \textbf{PPO\_CDA\_HRO}. Including these, we evaluated a total of six policy models in our RLHF experiments:

\begin{itemize}
    \item \textbf{OpenLM}: The unaligned base model with no fine-tuning.
    \item \textbf{SFT}: The supervised fine-tuned model trained on instruction-following data.
    \item \textbf{PPO\_HRO}: Fine-tuned via PPO using the baseline reward model HRO.
    \item \textbf{ODIN}: RLHF policy trained with the ODIN reward model reimplemented on OpenLLaMA-3B.
    \item \textbf{PPO\_CDA\_OpenLM}: Fine-tuned via PPO using our counterfactual reward model CDA\_OpenLM.
    \item \textbf{PPO\_CDA\_HRO}: Fine-tuned via PPO using our counterfactual reward model CDA\_HRO.
\end{itemize}

All PPO-based models were trained using DeepSpeed-Chat~\citep{yao2023dschat} with bfloat16 precision and distributed optimization across multiple GPUs. Unless otherwise noted, default PPO hyperparameters were used. Each model was initialized from the same SFT checkpoint and trained with its respective reward model.

\paragraph{Evaluation.}
We evaluate all PPO-based policies using AlpacaEval~\citep{dubois2025lengthcontrolledalpacaevalsimpleway}, reporting relative win rates against the \texttt{LLaMA-2-7B-chat-hf} reference model~\citep{touvron2023llama}. All models are assessed under identical conditions. See~\Cref{appendix:alpacaeval} for the complete evaluation protocol.

\subsection{ODIN Fine-Tuning Details}
\label{appendix:odin}

\paragraph{Reward model training.}
We reimplemented the ODIN reward model on the \texttt{open\_llama\_3b} backbone using Hugging Face's \texttt{Trainer} API. To ensure fair and architecture-consistent comparisons, we used the same base model (\texttt{open\_llama\_3b}) as our own reward models. The model was fine-tuned for 3 epochs with a sequence length cap of 512 tokens, using Adam8bit (\texttt{bitsandbytes}) with a learning rate of $2 \times 10^{-5}$ and weight decay of 0.01. FP16 precision and gradient checkpointing were used to optimize memory efficiency. Padding was handled via the EOS token, and the loss ignored padded labels using label pad token id = -100. Evaluation and checkpointing occurred every 50 steps.

% \begin{itemize}[itemsep=2pt, topsep=2pt]
%     \item Model: \texttt{open\_llama\_3b}
%     \item Tokenizer padding token: EOS
%     \item Max sequence length: 512 tokens
%     \item Optimizer: Adam8bit (bitsandbytes)
%     \item Learning rate: $2 \times 10^{-5}$
%     \item Weight decay: 0.01
%     \item Warmup ratio: 0.03
%     \item Batch size: 8 (per device)
%     \item Epochs: 3
%     \item Evaluation / Save steps: every 50 steps
%     \item Precision: \texttt{fp16}
%     \item Gradient checkpointing: Enabled
%     \item Data collator: \texttt{DataCollatorForSeq2Seq} with \texttt{label\_pad\_token\_id = -100}
% \end{itemize}
\begin{table}[ht]
\centering
\begin{tabular}{p{0.25\linewidth} p{0.38\linewidth}}
\hline
\textbf{Hyperparameter} & \textbf{Value} \\
    \hline
    Model & \texttt{open\_llama\_3b} \\
    Tokenizer padding token & EOS \\
    Maximum sequence length & 1024 tokens \\
    Optimizer & Adam8bit (\texttt{bitsandbytes}) \\
    Learning rate & $2 \times 10^{-5}$ \\
    Weight decay & 0.01 \\
    Batch size (per device) & 8 \\
    Epochs & 3 \\
    Evaluation / Save steps & Every 50 steps \\
    Precision & FP16 \\
    Gradient clipping & 1.0 \\
    Gradient checkpointing & Enabled \\
    Data collator & \texttt{DataCollatorForSeq2Seq} \\
    Label pad token ID & \texttt{-100} (ignored in loss) \\
    \hline
\end{tabular}
\caption{Training configuration for ODIN reward model fine-tuning.}
\label{tab:odin_finetuning}
\end{table}

\paragraph{PPO training.}
For the PPO phase, we used the \texttt{PPOTrainer} module from the TRL library to train a policy model using the ODIN reward signal. Training was conducted over 768 PPO steps with 4 epochs per batch and a mini-batch size of 4. The optimizer was Adam (also from \texttt{bitsandbytes}), with a learning rate of $5 \times 10^{-6}$. The KL penalty coefficient was set to 0.1, with a target KL divergence of 6 to maintain response diversity. The maximum response length was 128 tokens. Training used FP16 precision and gradient checkpointing, with logging performed at every step.

% \begin{itemize}[itemsep=2pt, topsep=2pt]
%     \item Model: \texttt{open\_llama\_3b}
%     \item Tokenizer padding token: EOS
%     \item Max response length: 128 tokens
%     \item Optimizer: Adam (bitsandbytes)
%     \item Learning rate: $5 \times 10^{-6}$
%     \item Batch size (per GPU): 1
%     \item PPO epochs per batch: 4
%     \item Mini-batch size: 4
%     \item KL penalty coefficient: 0.1
%     \item Target KL: 6
%     \item PPO steps: 768
%     \item Precision: \texttt{fp16}
%     \item Gradient checkpointing: Enabled
%     \item Log interval: Every 1 step
% \end{itemize}
\begin{table}[ht]
\centering
\begin{tabular}{p{0.25\linewidth} p{0.38\linewidth}}
\hline
\textbf{Hyperparameter} & \textbf{Value} \\
    \hline
    Model & \texttt{open\_llama\_3b} \\
    Tokenizer padding token & EOS \\
    Maximum response length & 1024 tokens \\
    Optimizer & Adam (\texttt{bitsandbytes}) \\
    Learning rate & $5 \times 10^{-7}$ \\
    Batch size (per GPU) & 1 \\
    PPO epochs per batch & 4 \\
    Mini-batch size & 4 \\
    KL penalty coefficient & 0.1 \\
    Target KL & 6 \\
    PPO steps & 768 \\
    Precision & FP16 \\
    Gradient checkpointing & Enabled \\
    Log interval & Every 1 step \\
    \hline
\end{tabular}
\caption{PPO training configuration using the ODIN reward model.}
\label{tab:odin_ppo}
\end{table}

\paragraph{Evaluation.}
The ODIN reward model was evaluated using the same protocol as our proposed reward models. During training, we monitored pairwise accuracy on the validation set. For final evaluation, the model was assessed on RewardBench and length-controlled accuracy metrics, as described in~\Cref{appendix:alpacaeval}. The PPO-trained policy using the ODIN reward model was evaluated using AlpacaEval~\citep{dubois2025lengthcontrolledalpacaevalsimpleway}, reporting relative win rates against the \texttt{LLaMA-2-7B-chat-hf} reference model under the same length-controlled setup as our other PPO-based policies.

\section{Flip Ratio Computation and Thresholding}
\label{appendix:flip}

We recall the definition of the flip ratio \(F(A, B)\), introduced in the main text:
\[
F(A, B) =
\frac{
\# \, \text{of comparisons where preference flipped}
}{
\text{Total number of counterfactual comparisons}
}
\]
This metric quantifies how frequently a model's original preference is reversed when exposed to length-controlled counterfactuals with matched semantic content.

We classify a pair as \textit{length-biased} if \(F(A, B) > 0.5\), i.e., when a majority of the content-fixed variants lead to a reversal in the model’s decision. This conservative threshold ensures that only cases with consistent flip behavior are flagged, thereby mitigating the influence of noise or isolated inconsistencies.

\begin{figure}[t]
  \centering
  \includegraphics[width=0.65\linewidth]{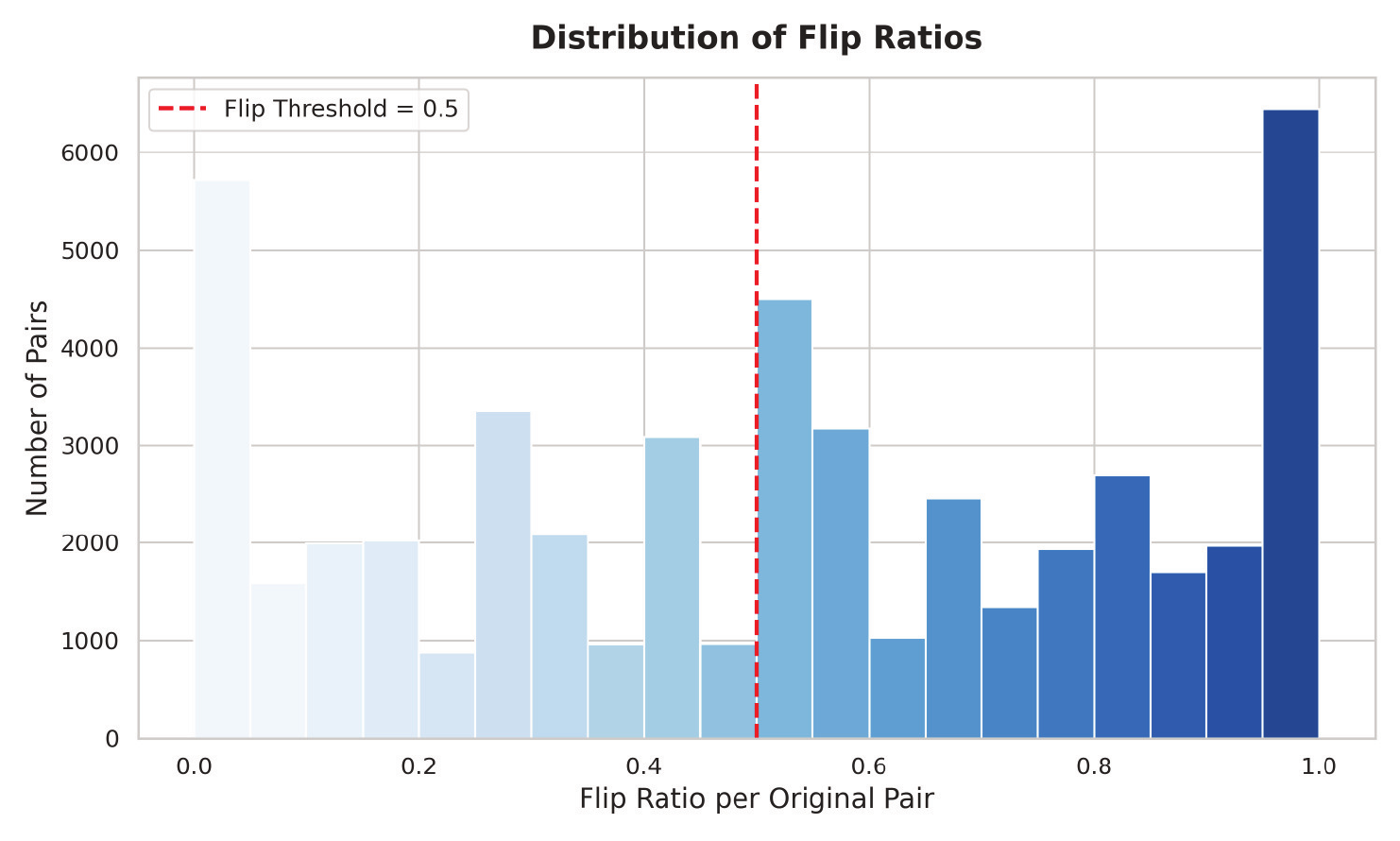}
  \caption{Distribution of flip ratios per original preference pair. A flip ratio of 1.0 indicates consistent preference reversals, implying strong structural length bias.}
  \label{fig:flip-ratio-dist}
\end{figure}

As shown in \Cref{fig:flip-ratio-dist}, the distribution of flip ratios exhibits a clear bimodal structure, with prominent peaks near $0.0$ and $1.0$. This suggests that most model decisions are either highly robust or highly sensitive to changes in response length. The choice of $0.5$ as the decision threshold is motivated by the trough between the two peaks. It separates stable decisions from structurally biased ones, enabling a binary classification scheme that is both scalable and aligned with human-intuitive diagnostic rules.

\begin{itemize}
    \item \textbf{Robust Region ($F \approx 0.0$):} Pairs for which the model’s preference remains unchanged across all counterfactuals, indicating stability with respect to length variation.
    \item \textbf{Biased Region ($F \approx 1.0$):} Pairs where every length-modified variant leads to a flipped preference, revealing strong susceptibility to verbosity.
    \item \textbf{Unstable Region ($F \approx 0.5$):} Pairs with inconsistent behavior, where the model is sensitive to minor, non-semantic length variations. These may reflect borderline or stylistically-driven decisions.
\end{itemize}

This distribution validates the utility of the flip ratio not only as a diagnostic signal but also as a filtering mechanism for training or evaluation data where verbosity should not dominate reward assignment.

\section{Evaluation Results}
\label{appendix:alpacaeval}
% This paper formally describes evaluation metrics used
% and explains the motivation for choosing these met-
% rics (yes/partial/no) => 데이터셋 설명 + 이유 추가까지 하기
% bless hyeonji..

% \paragraph{Motivation for RewardBench Evaluation.}
\subsection{Reward Model Evaluations}
While mitigating length bias is essential for aligning reward models with human preferences, prior work has consistently shown a trade-off between bias reduction and overall performance~\citep{shen2023loose, chen2024odin, huang2024post}. That is, reward models that suppress verbosity often suffer from diminished ability to distinguish genuinely helpful responses—particularly in settings where length correlates with informativeness. Our objective, however, is not only to reduce length bias but also to maintain or improve general reward model performance. 

To assess this dual goal, we employ the \textbf{RewardBench} benchmark suite~\citep{lambert2024rewardbench, malik2025rewardbench2advancingreward}, a comprehensive evaluation framework that tests reward models across diverse dimensions including open-domain helpfulness, safety, and reasoning. Unlike length-controlled diagnostics, RewardBench allows us to measure whether our counterfactual fine-tuning method preserves alignment quality while mitigating spurious length effects. This evaluation ensures that any reduction in length bias does not come at the cost of degraded general alignment. As a diagnostic for length bias, we also report \textbf{Length-Controlled (LC) Accuracy}, which directly quantifies a reward model's tendency to prefer more concise responses when appropriate. This metric serves as a targeted measure of verbosity bias mitigation.

\begin{table}[t]
\centering
\small
\begin{tabular}{lcccccc}
\toprule
\textbf{Model} & \textbf{Iter} & \textbf{Chat} & \textbf{Chat Hard} & \textbf{Safety} & \textbf{Reasoning} & \textbf{Avg} \\
\midrule
\multirow{4}{*}{HRO}
    & 1 & 0.718 & 0.485 & 0.334 & 0.420 & 0.486 \\
    & 2 & 0.718 & 0.485 & 0.334 & 0.420 & 0.486 \\
    & 3 & 0.718 & 0.485 & 0.334 & 0.420 & 0.486 \\
    & \textbf{Mean} & \textbf{0.718} & \textbf{0.485} & \textbf{0.334} & \textbf{0.420} & \textbf{0.486} \\
\midrule
\multirow{4}{*}{ODIN}
    & 1 & 0.499 & 0.487 & 0.514 & 0.485 & 0.496 \\
    & 2 & 0.499 & 0.487 & 0.514 & 0.485 & 0.496 \\
    & 3 & 0.499 & 0.487 & 0.514 & 0.485 & 0.496 \\
    & \textbf{Mean} & \textbf{0.499} & \textbf{0.487} & \textbf{0.514} & \textbf{0.485} & \textbf{0.496} \\
\midrule
\multirow{4}{*}{CDA\_OpenLM*}
    & 1 & 0.508 & 0.458 & 0.484 & 0.460 & 0.478 \\
    & 2 & 0.444 & 0.496 & 0.507 & 0.506 & 0.488 \\
    & 3 & 0.444 & 0.524 & 0.520 & 0.479 & 0.492 \\
    & \textbf{Mean} & \textbf{0.466} & \textbf{0.493} & \textbf{0.504} & \textbf{0.482} & \textbf{0.486} \\
\midrule
\multirow{4}{*}{CDA\_LoRA*}
    & 1 & 0.732 & 0.496 & 0.332 & 0.427 & 0.497 \\
    & 2 & 0.732 & 0.496 & 0.332 & 0.427 & 0.497 \\
    & 3 & 0.732 & 0.496 & 0.332 & 0.427 & 0.497 \\
    & \textbf{Mean} & \textbf{0.732} & \textbf{0.496} & \textbf{0.332} & \textbf{0.427} & \textbf{0.497} \\
\midrule
\multirow{4}{*}{CDA\_HRO*}
    & 1 & 0.453 & 0.509 & 0.568 & 0.507 & 0.509 \\
    & 2 & 0.525 & 0.500 & 0.496 & 0.496 & 0.504 \\
    & 3 & 0.494 & 0.520 & 0.523 & 0.481 & 0.505 \\
    & \textbf{Mean} & \textbf{0.491} & \textbf{0.510} & \textbf{0.529} & \textbf{0.495} & \textbf{0.506} \\
\bottomrule
\end{tabular}
\caption{Results for each of the 3 independent runs and their mean on RewardBench-1.}
\label{tab:rewardbench1_iterations}
\end{table}

\paragraph{RewardBench-1.}

We evaluate variants of the reward model using \texttt{RewardBench-1}, a composite benchmark designed for diagnosing various alignment shortcomings of reward models. RewardBench-1 consists of four major subsets derived from established public test collections:

\begin{itemize}
  \item \textbf{Chat}: Easy chat comparisons from \texttt{alpacaeval‑easy}, \texttt{alpacaeval‑length}, \texttt{alpacaeval‑hard}, \texttt{mt‑bench‑easy}, and \texttt{mt‑bench‑medium}, covering fluent dialogues with superficial stylistic or length perturbations.
  \item \textbf{Chat‐Hard}: More challenging variants including \texttt{mt‑bench‑hard}, \texttt{llmbar‑natural}, and various adversarial instruction prompts (\texttt{llmbar‑adver‑*}) with high semantic overlap but structural differences.
  \item \textbf{Safety}: Safety‐sensitive tests from \texttt{refusals‑dangerous}, \texttt{refusals‑offensive}, \texttt{xstest‑should‑refuse}, \texttt{xstest‑should‑respond}, and \texttt{do not answer}, evaluating the model on downstream refusal and toxicity scenarios.
  \item \textbf{Reasoning}: Reasoning and code‐based subsets (\texttt{math‑prm}, \texttt{hep‑cpp}, \texttt{hep‑go}, \texttt{hep‑java}, \texttt{hep‑js}, \texttt{hep‑python}, \texttt{hep‑rust}) testing logical correctness and inference capability.
\end{itemize}

Each subset is evaluated via pairwise accuracy (i.e., whether the reward model assigns a higher score to the gold response than the distractor). Subset scores are computed via per‑prompt weighted averaging, and the overall RewardBench score is the average across subset scores. ~\Cref{tab:rewardbench1_iterations} presents the performance of each reward model on RewardBench-1 across independent fine-tuning runs.

\paragraph{RewardBench-2.}

To assess the generalization capabilities of our reward models across a broader range of human preferences, beyond just verbosity-controlled scenarios, we further evaluate on \textsc{RewardBench-2}. This follow-up benchmark includes a set of completely unseen human prompts and introduces more complex, real-world multi-choice formats, providing a more challenging evaluation of model performance across diverse preference rankings and decision-making tasks.

The benchmark comprises six domains sourced primarily from novel human-written prompts via the WildChat pipeline, with rigorous cleaning and filtering to avoid overlap with existing evaluation datasets~\citep{malik2025rewardbench2advancingreward}: 
\emph{Factuality}, \emph{Precise Instruction Following}, \emph{Math}, \emph{Safety}, \emph{Focus}, and \emph{Ties}.  
Completion sets are generated by multiple LLMs and filtered through combinations of manual verification, multi‑LLM judgment, and rule-based constraint verifiers:

\begin{itemize}
  \item \textbf{Factuality}: 475 prompts labeled based on multi-LLM agreement to detect hallucinations.
  \item \textbf{Precise Instruction Following}: 160 prompts testing adherence to exact constraints (e.g. “no letter u”) via verifier functions.
  \item \textbf{Math}: 183 math-related prompts from human users, judged by majority voting.
  \item \textbf{Safety}: 450 prompts (including from CoCoNot), evaluated using LLM‑judging and human rubrics.
  \item \textbf{Focus}: 495 prompts that test relevance and topicality using system-prompt variation.
  \item \textbf{Ties}: 102 prompts with multiple valid answers, evaluated by weighted accuracy scoring across correct and incorrect answers.
\end{itemize}

\begin{table}[t]
\centering
\small
\begin{tabular}{lccccccc}
\toprule
\textbf{Model} & \textbf{Iter} & \textbf{Factuality} & \textbf{PIF} & \textbf{Math} & \textbf{Safety} & \textbf{Focus} & \textbf{Avg} \\
\midrule
\multirow{4}{*}{HRO}
    & 1 & 0.364 & 0.275 & 0.350 & 0.240 & 0.238 & 0.250 \\
    & 2 & 0.364 & 0.275 & 0.350 & 0.240 & 0.238 & 0.250 \\
    & 3 & 0.364 & 0.275 & 0.350 & 0.240 & 0.238 & 0.250 \\
    & \textbf{Mean} & \textbf{0.364} & \textbf{0.275} & \textbf{0.350} & \textbf{0.240} & \textbf{0.238} & \textbf{0.250} \\
\midrule
\multirow{4}{*}{ODIN}
    & 1 & 0.301 & 0.263 & 0.230 & 0.154 & 0.147 & 0.219 \\
    & 2 & 0.301 & 0.263 & 0.230 & 0.154 & 0.147 & 0.219 \\
    & 3 & 0.301 & 0.263 & 0.230 & 0.154 & 0.147 & 0.219 \\
    & \textbf{Mean} & \textbf{0.301} & \textbf{0.263} & \textbf{0.230} & \textbf{0.154} & \textbf{0.147} & \textbf{0.219} \\
\midrule
\multirow{4}{*}{CDA\_OpenLM*}
    & 1 & 0.412 & 0.244 & 0.311 & 0.264 & 0.139 & 0.274 \\
    & 2 & 0.430 & 0.271 & 0.311 & 0.271 & 0.137 & 0.284 \\
    & 3 & 0.406 & 0.256 & 0.311 & 0.276 & 0.124 & 0.275 \\
    & \textbf{Mean} & \textbf{0.416} & \textbf{0.257} & \textbf{0.311} & \textbf{0.270} & \textbf{0.133} & \textbf{0.278} \\
\midrule
\multirow{4}{*}{CDA\_LoRA*}
    & 2 & 0.361 & 0.244 & 0.336 & 0.267 & 0.232 & 0.288 \\
    & 3 & 0.361 & 0.244 & 0.336 & 0.267 & 0.232 & 0.288 \\
    & 1 & 0.361 & 0.244 & 0.336 & 0.267 & 0.232 & 0.288 \\
    & \textbf{Mean} & \textbf{0.361} & \textbf{0.244} & \textbf{0.336} & \textbf{0.267} & \textbf{0.232} & \textbf{0.288} \\
\midrule
\multirow{4}{*}{CDA\_HRO*}
    & 1 & 0.461 & 0.194 & 0.246 & 0.280 & 0.198 & 0.276 \\
    & 2 & 0.470 & 0.206 & 0.246 & 0.272 & 0.202 & 0.279 \\
    & 3 & 0.451 & 0.191 & 0.241 & 0.292 & 0.197 & 0.274 \\
    & \textbf{Mean} & \textbf{0.461} & \textbf{0.197} & \textbf{0.244} & \textbf{0.281} & \textbf{0.199} & \textbf{0.276} \\
\bottomrule
\end{tabular}
\caption{Results for each of the 3 independent runs and their mean on RewardBench-2.}
\label{tab:rewardbench2_iterations}
\end{table}

Each prompt is paired with four candidate completions (one chosen, three rejected) except in \textit{Ties}, where multiple correct completions exist. Models must select the single chosen completion to score as correct; the Ties domain uses a custom weighted scoring scheme to reward discrimination among valid answers~\citep{malik2025rewardbench2advancingreward}. 

For consistency with our length bias diagnosis and mitigation methodology, we exclude tie cases from \textsc{RewardBench-2} in this report. In our approach, length bias is diagnosed when a model’s preference for a longer response is driven by verbosity rather than content quality. Since tie cases involve multiple valid completions with no clear preference, they do not provide a meaningful test for length bias. Including ties could introduce noise into our evaluation, as they do not reflect a preference influenced by response length, but rather an equal ranking of multiple answers. By excluding these cases, we ensure that our evaluation remains focused on assessing the impact of length bias on model preferences.

\textsc{RewardBench-2} is designed with a robust evaluation structure to assess model performance across diverse domains. The final performance score is the unweighted average of per-domain accuracies. The inclusion of a best-of-4 format, unseen prompts, and expanded domains allows \textsc{RewardBench-2} to evaluate model performance across a wider range of areas, which were not covered in \textsc{RewardBench-1}. This enables a more comprehensive assessment of reward models, including performance in previously unexplored domains. Results across independent runs are shown in \Cref{tab:rewardbench2_iterations}..

\paragraph{Length-Controlled Accuracy (LC Accuracy)}

To evaluate whether reward models correctly prioritize semantic quality over surface features like verbosity, we introduce \textbf{Length-Controlled Accuracy (LC Accuracy)} as a metric for length bias mitigation.

This metric is computed using a filtered subset of the Chatbot Arena pairwise preference dataset~\citep{chiang2024chatbot}, which offers a diverse set of prompts and responses with varying lengths, making it particularly well-suited for detecting length bias in model evaluations. Specifically, we select comparison pairs where the preferred (winning) response is \emph{shorter} than the rejected one, with a minimum difference of two token-length bins (e.g., short vs long, medium vs very long, etc.). These conditions ensure that the comparison isolates length preference rather than content or formatting artifacts. We then calculate the accuracy of each reward model in correctly selecting the shorter response as the winner. The iteration performance is shown in \Cref{tab:chatbotarena_iterations}.

\paragraph{Reward models evaluation conclusion.}
Our CDA models (\texttt{CDA\_OpenLM}, \texttt{CDA\_LoRA}, \texttt{CDA\_HRO}) achieve overall RewardBench performance comparable to the baseline models (\texttt{HRO}, \texttt{ODIN}), indicating that counterfactual augmentation does not degrade general reward quality. At the same time, CDA-based reward models show more stable category-wise behavior across RewardBench subsets compared to baseline models, which often exhibit uneven performance across tasks. Moreover, they substantially improve Length-Controlled Accuracy, with \texttt{CDA\_OpenLM} and \texttt{CDA\_HRO} outperforming all baselines by a large margin. This demonstrates that our method effectively mitigates length bias. Together, these results show that our approach improves robustness to verbosity while maintaining competitive overall accuracy—successfully avoiding the performance–bias trade-off observed in prior methods.

\begin{table}[t]
\centering
\small
\begin{tabular}{p{1.2cm}p{1.2cm}p{1.2cm}p{2.0cm}p{1.8cm}p{1.8cm}}
\toprule
\textbf{Iter} & \textbf{HRO} & \textbf{ODIN} & \textbf{CDA\_OpenLM*} & \textbf{CDA\_LoRA*} & \textbf{CDA\_HRO*} \\
\midrule
1 & 0.249 & 0.473 & 0.487 & 0.249 & 0.478 \\
2 & 0.249 & 0.436 & 0.530 & 0.249 & 0.498 \\
3 & 0.249 & 0.480 & 0.506 & 0.249 & 0.502 \\
\textbf{Mean} & \textbf{0.249} & \textbf{0.463} & \textbf{0.508} & \textbf{0.249} & \textbf{0.493} \\
\bottomrule
\end{tabular}
\caption{Length-controlled accuracy from Chatbot Arena for each of the 3 independent runs and their mean.}
\label{tab:chatbotarena_iterations}
\end{table}

\subsection{PPO Evaluations}
\paragraph{AlpacaEval.}
\textbf{AlpacaEval}~\citep{dubois2025lengthcontrolledalpacaevalsimpleway} is a comprehensive large-scale evaluation benchmark specifically designed to assess instruction-following models through pairwise preference comparisons. In this evaluation protocol, each model generates responses to a shared set of prompts, and the quality of these responses is assessed via pairwise comparisons. The evaluation is carried out using a reference judge model—typically the \texttt{LLaMA-2-7B-chat} model~\citep{touvron2023llama}. This reference model is tasked with determining which of the two generated responses better adheres to key instruction-following criteria such as helpfulness, relevance, and coherence. These criteria are critical for evaluating the effectiveness of the models in generating human-like, contextually appropriate responses.

Given that our base policy model, OpenLLaMA-3B, is derived from LLaMA-1, we chose to use \texttt{LLaMA-2-7B-chat} as our reference model for comparison. This choice allows us to leverage both the upgraded version (LLaMA-2 compared to LLaMA-1) and a larger model size (7B compared to 3B) to ensure a more robust and up-to-date reference judge. By using a more advanced model, we ensure that the comparison is made against a stronger benchmark, providing a more accurate reflection of our model's performance in handling complex instruction-following tasks and mitigating length bias.

AlpacaEval offers several variants, one of which is the \textit{Length-Controlled AlpacaEval} (LCAE) version. LCAE specifically focuses on cases where the shorter response is preferred, and it is designed to assess the models' ability to produce concise yet high-quality responses. This variant isolates scenarios where verbosity, or longer responses, should not be favored over brevity, providing a more controlled environment for evaluating the impact of length bias. In this context, the \textbf{LC Winrate} is the key metric used to assess the model's performance in LCAE. 

To calculate the \textbf{LC Winrate}, a logistic regression model is used, considering three factors: \textit{model identity}, \textit{output length}, and \textit{instruction difficulty}. The model identity determines whether the output is from the baseline or evaluated model, while the output length accounts for the length bias in traditional evaluations. Instruction difficulty captures variations in task complexity. The length term is normalized, and the LC Winrate is computed by adjusting for length differences, providing a more fair assessment of model performance by removing length-related biases. Refer to~\citep{dubois2025lengthcontrolledalpacaevalsimpleway} for more detailed explanations.

So, to see how our PPO models mitigated length bias, we analyzed LC Winrate, Winrate, and average token length. Iteration performance can be found in~\Cref{tab:alpacaeval_iterations}.

% alpaca 데이터셋 설명
% iteration별 결과
\begin{table}[t]
\centering
\small
\begin{tabular}{lcccc}
\toprule
\textbf{Model} & \textbf{Iter} & \textbf{LC Winrate} & \textbf{Winrate} & \textbf{Avg. length} \\
\midrule
\multirow{4}{*}{SFT}
    & 1 & 21.07 & 31.55 & 2058 \\
    & 2 & 14.43 & 23.85 & 2071 \\
    & 3 & 13.42 & 21.74 & 2054 \\
    & \textbf{Mean} & \textbf{16.97} & \textbf{25.71} & \textbf{2061} \\
\midrule
\multirow{4}{*}{PPO\_HRO}
    & 1 & 19.71 & 28.57 & 2052 \\
    & 2 & 16.90 & 27.70 & 2048 \\
    & 3 & 20.31 & 29.07 & 2043 \\
    & \textbf{Mean} & \textbf{18.97} & \textbf{28.45} & \textbf{2048} \\
\midrule
\multirow{4}{*}{OpenLM}
    & 1 & 9.02 & 10.43 & 1379 \\
    & 2 & 7.72 & 9.69 & 1399 \\
    & 3 & 8.68 & 9.69 & 1378 \\
    & \textbf{Mean} & \textbf{8.47} & \textbf{9.94} & \textbf{1385} \\
\midrule
\multirow{4}{*}{ODIN}
    & 1 & 12.19 & 11.30 & 1023 \\
    & 2 & 12.47 & 11.55 & 1008 \\
    & 3 & 11.90 & 11.18 & 1047 \\
    & \textbf{Mean} & \textbf{12.19} & \textbf{11.34} & \textbf{1026} \\
\midrule
\multirow{4}{*}{PPO\_CDA\_HRO*}
    & 1 & 37.65 & 32.80 & 1116 \\
    & 2 & 36.00 & 29.94 & 1052 \\
    & 3 & 34.54 & 29.32 & 1049 \\
    & \textbf{Mean} & \textbf{36.06} & \textbf{30.69} & \textbf{1072} \\
\midrule
\multirow{4}{*}{PPO\_CDA\_OpenLM*}
    & 1 & 36.91 & 32.67 & 1139 \\
    & 2 & 39.37 & 34.29 & 1108 \\
    & 3 & 35.25 & 30.68 & 1108 \\
    & \textbf{Mean} & \textbf{37.18} & \textbf{32.55} & \textbf{1118} \\
\bottomrule
\end{tabular}
\caption{Length controlled winrate for each of the 3 independent runs and their mean on AlpacaEval.}
\label{tab:alpacaeval_iterations}
\end{table}

\paragraph{PPO models evaluation conclusion.}
The results from the AlpacaEval evaluation reveal that our PPO-trained models—\texttt{PPO\_CDA\_HRO} and \texttt{PPO\_CDA\_OpenLM}—significantly outperform other models, including the baseline models and SFT, in terms of Length-Controlled Winrate (LC Winrate). While the SFT model and baseline models such as ODIN and OpenLM show consistently low performance in length-controlled scenarios, our PPO models consistently achieve higher LC winrates, demonstrating a strong ability to prioritize concise responses without sacrificing performance. 

In particular, \texttt{PPO\_CDA\_OpenLM} and \texttt{PPO\_CDA\_HRO} models not only outperform baseline models on LC Winrate but also maintain competitive performance in standard evaluation metrics, such as overall Winrate. This showcases the effectiveness of our approach in mitigating length bias while preserving model quality across various tasks. The lower average response length for the PPO models, especially compared to SFT and baseline models, highlights their ability to produce shorter yet high-quality responses. These results confirm that the PPO-trained models successfully navigate the traditional trade-off between bias reduction and overall performance, establishing them as robust and length-efficient models.

\section{Evaluation with More Recent Baseline Models}
\label{appendix:recent-baselines}
To verify that the improvements observed in our main evaluation do not depend on the specific choice of judge model, we additionally evaluate all PPO variants using two more recent and substantially stronger reference models: \texttt{Meta-LLaMA-3-8B-Instruct} and \texttt{Meta-LLaMA-3.1-8B-Instruct-Turbo}. While our primary experiments follow the standard AlpacaEval protocol, which uses \texttt{LLaMA-2-7B-chat} for consistency with prior work, these supplementary results allow us to confirm that our conclusions hold under modern, higher-capacity judges.

\Cref{tab:alpacaeval_iterations_llama3} and \Cref{tab:alpacaeval_iterations_llama3_turbo} summarize LC winrate, overall winrate, and average output length across three independent runs under the two new judges. As expected, absolute winrates decrease due to the stronger decision boundaries imposed by LLaMA-3–family models. However, the relative ordering of methods remains consistent with the findings reported in the main paper.

Across both judge models, our counterfactually trained PPO policies---\texttt{PPO\_CDA\_OpenLM} and \texttt{PPO\_CDA\_HRO}---retain a substantial advantage in Length-Controlled winrate over all baselines. The margin of improvement over \texttt{SFT}, \texttt{PPO\_HRO}, \texttt{OpenLM}, and \texttt{ODIN} is even more pronounced under the LLaMA-3 judges, indicating that stronger evaluators more clearly distinguish concise, semantically grounded responses from verbose outputs. Moreover, CDA-based models maintain competitive overall winrate despite producing significantly shorter responses, consistent with the behavior observed under the LLaMA-2 judge.

Taken together, these supplementary evaluations demonstrate that the benefits of counterfactual reward model fine-tuning are not tied to a particular evaluation setup. Even under modern, high-capacity judge models, CDA-trained PPO policies consistently achieve the highest robustness to verbosity while preserving general response quality, further validating the generality of our approach.

\begin{table}[ht]
\centering
\small
\begin{tabular}{lcccc}
\toprule
\textbf{Model} & \textbf{Iter} & \textbf{LC Winrate} & \textbf{Winrate} & \textbf{Avg. length} \\
\midrule
\multirow{4}{*}{SFT}
    & 1 & 1.08 & 2.73 & 2063 \\
    & 2 & 2.25 & 3.18 & 2071 \\
    & 3 & 1.33 & 2.48 & 2054 \\
    & \textbf{Mean} & \textbf{1.72} & \textbf{2.90} & \textbf{2062} \\
\midrule
\multirow{4}{*}{PPO\_HRO}
    & 1 & 1.96 & 3.48 & 2043 \\
    & 2 & 2.19 & 3.73 & 2048 \\
    & 3 & 1.75 & 2.86 & 2064 \\
    & \textbf{Mean} & \textbf{1.97} & \textbf{3.36} & \textbf{2051} \\
\midrule
\multirow{4}{*}{OpenLM}
    & 1 & 0.95 & 1.24 & 1378 \\
    & 2 & 0.57 & 0.75 & 1379 \\
    & 3 & 0.42 & 0.62 & 1399 \\
    & \textbf{Mean} & \textbf{0.65} & \textbf{0.87} & \textbf{1385} \\
\midrule
\multirow{4}{*}{ODIN}
    & 1 & 0.82 & 0.75 & 1047 \\
    & 2 & 0.84 & 0.75 & 1008 \\
    & 3 & 1.02 & 0.87 & 1023 \\
    & \textbf{Mean} & \textbf{0.89} & \textbf{0.79} & \textbf{1026} \\
\midrule
\multirow{4}{*}{PPO\_CDA\_HRO*}
    & 1 & 5.29 & 3.60 & 1049 \\
    & 2 & 6.65 & 4.84 & 1114 \\
    & 3 & 6.80 & 4.91 & 1115 \\
    & \textbf{Mean} & \textbf{6.25} & \textbf{4.45} & \textbf{1093} \\
\midrule
\multirow{4}{*}{PPO\_CDA\_OpenLM*}
    & 1 & 6.55 & 4.66 & 1108 \\
    & 2 & 6.10 & 4.41 & 1121 \\
    & 3 & 5.84 & 4.22 & 1124 \\
    & \textbf{Mean} & \textbf{6.16} & \textbf{4.43} & \textbf{1118} \\
\bottomrule
\end{tabular}
\caption{AlpacaEval results evaluated using \texttt{Meta-LLaMA-3-8B-Instruct} as the judge model.} 
\label{tab:alpacaeval_iterations_llama3}
\end{table}

\begin{table}[H]
\centering
\small
\begin{tabular}{lcccc}
\toprule
\textbf{Model} & \textbf{Iter} & \textbf{LC Winrate} & \textbf{Winrate} & \textbf{Avg. length} \\
\midrule
\multirow{4}{*}{SFT}
    & 1 & 4.35 & 5.37 & 2063 \\
    & 2 & 5.54 & 6.58 & 2071 \\
    & 3 & 4.53 & 5.59 & 2054 \\
    & \textbf{Mean} & \textbf{4.81} & \textbf{5.85} & \textbf{2062} \\
\midrule
\multirow{4}{*}{PPO\_HRO}
    & 1 & 7.16 & 8.82 & 2043 \\
    & 2 & 7.16 & 8.20 & 2048 \\
    & 3 & 6.93 & 8.07 & 2064 \\
    & \textbf{Mean} & \textbf{7.08} & \textbf{8.36} & \textbf{2051} \\
\midrule
\multirow{4}{*}{OpenLM}
    & 1 & 2.00 & 2.11 & 1378 \\
    & 2 & 1.60 & 1.74 & 1379 \\
    & 3 & 1.46 & 1.61 & 1399 \\
    & \textbf{Mean} & \textbf{1.69} & \textbf{1.82} & \textbf{1385} \\
\midrule
\multirow{4}{*}{ODIN}
    & 1 & 1.86 & 1.49 & 1047 \\
    & 2 & 2.05 & 1.61 & 1008 \\
    & 3 & 2.36 & 1.86 & 1023 \\
    & \textbf{Mean} & \textbf{2.09} & \textbf{1.65} & \textbf{1026} \\
\midrule
\multirow{4}{*}{PPO\_CDA\_HRO*}
    & 1 & 13.27 & 9.81 & 1114 \\
    & 2 & 12.87 & 10.12 & 1115 \\
    & 3 & 11.47 & 8.20 & 1049 \\
    & \textbf{Mean} & \textbf{12.54} & \textbf{9.38} & \textbf{1093} \\
\midrule
\multirow{4}{*}{PPO\_CDA\_OpenLM*}
    & 1 & 12.36 & 9.32 & 1108 \\
    & 2 & 12.50 & 9.38 & 1108 \\
    & 3 & 12.72 & 9.32 & 1124 \\
    & \textbf{Mean} & \textbf{12.53} & \textbf{9.34} & \textbf{1118} \\
\bottomrule
\end{tabular}
\caption{AlpacaEval results evaluated using \texttt{Meta-LLaMA-3.1-8B-Instruct-Turbo} as the judge model.}
\label{tab:alpacaeval_iterations_llama3_turbo}
\end{table}

\end{document}